# Bounded Recursive Self-Improvement

E. Nivel,[1] K. R. Thórisson,[1,6] B. R. Steunebrink,[5]
H. Dindo,[2] G. Pezzulo,[4] M. Rodriguez,[3] C. Hernandez,[3]
D. Ognibene,[4] J. Schmidhuber,[5] R. Sanz,[3]
H. P. Helgason,[1] A. Chella[2] & G. K. Jonsson[7]

[1] Reykjavik University / CADIA
[2] Universita degli studi di Palermo / DINFO
[3] Universidad Politecnica de Madrid / ASLAB
[4] Consiglio Nazionale delle Ricerche / ISTC
[5] The Swiss AI Lab IDSIA, USI & SUPSI
[6] Icelandic Institute for Intelligent Machines
[7] University of Iceland / HBL

# Bounded Recursive Self-Improvement

E. Nivel,[1] K. R. Thórisson,[1,6] B. R. Steunebrink,[5]
H. Dindo,[2] G. Pezzulo,[4] M. Rodriguez,[3] C. Hernandez,[3]
D. Ognibene,[4] J. Schmidhuber,[5] R. Sanz,[3]
H. P. Helgason,[1] A. Chella[2] & G. K. Jonsson[7]

[1] Reykjavik University / CADIA
[2] Universita degli studi di Palermo / DINFO
[3] Universidad Politecnica de Madrid / ASLAB
[4] Consiglio Nazionale delle Ricerche / ISTC
[5] The Swiss AI Lab IDSIA, USI & SUPSI
[6] Icelandic Institute for Intelligent Machines
[7] University of Iceland / HBL

**Abstract.** Four principal features of autonomous control systems are left both unaddressed and unaddressable by present-day engineering methodologies: 1. The ability to operate effectively in environments that are only partially known beforehand at design time; 2. A level of generality that allows a system to re-assess and re-define the fulfillment of its mission in light of unexpected constraints or other unforeseen changes in the environment; 3. The ability to operate effectively in environments of significant complexity; and 4. The ability to degrade gracefully – how it can continue striving to achieve its main goals when resources become scarce, or in light of other expected or unexpected constraining factors that impede its progress. We describe new methodological and engineering principles for addressing these shortcomings, that we have used to design a machine that becomes increasingly better at behaving in underspecified circumstances, in a goal-directed way, on the job, by modeling itself and its environment as experience accumulates. Based on principles of autocatalysis, endogeny, and reflectivity, the work provides an architectural blueprint for constructing systems with high levels of operational autonomy in underspecified circumstances, starting from only a small amount of designer-specified code – a *seed*. Using a value-driven dynamic priority scheduling to control the parallel execution of a vast number of lines of reasoning, the system accumulates increasingly useful models of its experience, resulting in recursive self-improvement that can be autonomously sustained after the machine leaves the lab, within the boundaries imposed by its designers. A prototype system has been implemented and demonstrated to learn a complex real-world task – real-time multimodal dialogue with humans – by on-line observation. Our work presents solutions to several challenges that must be solved for achieving artificial general intelligence.

## 1 Introduction

Engineering is essentially to couple a system, an environment and a mission to meet a predefined set of requirements (Sanz, Matia & Galán, 2000). Present-day control architectures solve this problem only under the following assumption: The environment is well defined and so are the ways to achieve its mission. Given an ability to impart to a machine sufficient knowledge and resources for operating in a fully specified environment, systems can currently be built that meet task-related goals. What this approach doesn't address at all is: (a) How to build a system that is flexible and general enough to adapt when the environment changes from its initial specification, and to redefine accordingly its ways to fulfill its mission; (b) How to make a system cope with an environment the complexity of which allows only partial descriptions (i.e. when specifications become intractable); and (c) Since any physical system is resource-bounded, how to enable such systems to continue operation and degrade gracefully when their resources become scarce, or when adaptation to any other changes becomes necessary.

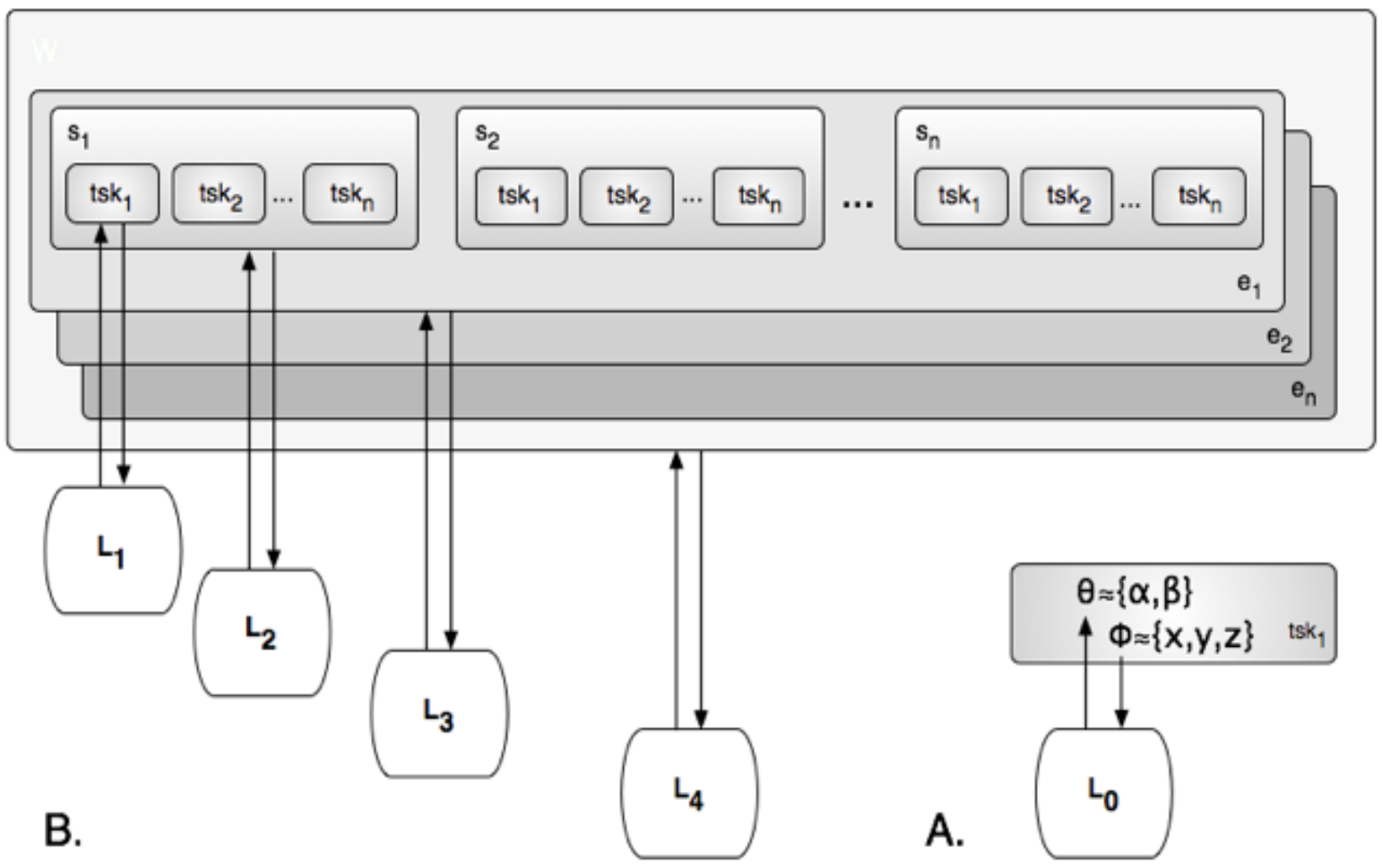

*Figure 1 – Scope and complexity of input to learning machines*

Intelligence is a technical means to the end of controlling a system's behavior so that it can survive adversity, be kept operating within its imposed and natural boundaries and still deliver the *value* it was intended to deliver. Our work concerns the creation of autonomous controllers that meet the unaddressed limitations of artificial intelligent systems, and this requires a re-evaluation of present engineering methodologies. First, to address the question of what a system can do *after it leaves the lab*, we must assume that the artificial general intelligent (AGI) systems of the future, to deserve the name, can learn complex tasks autonomously. Even tasks done routinely by 5-year olds, such as cleaning up the playroom, are beyond the capability of today's systems. As illustrated in Figure 1, current learning systems are limited to a handful of input and output parameters (Figure 1-A), typically on a single pre-defined task in a well-defined, unchanging environment. Let $tsk_i$ (Figure 1-B) refer to relatively non-trivial tasks such as assembling furniture and moving books, computers, tables, and chairs from one office to another, state-of-the-art machine-learning technique L0 is limited to learn only about a small subset of the various things that must be learned to achieve such a task. Being able to handle such a task in full, L1 in Figure 1-B is already more capable than most if not all advanced learning AI available today, being able to learn a single such complex task whose features were *unknown prior to the system's deployment*, even if it can only do so in one situation ($S_1$). L2, L3 and L4 take successive steps up the complexity ladder beyond that, being able to learn *a number* of complex tasks (L2), in *other situations* (L3), and in a wider *range of environments* and mission spaces (L4). Our work aims at toward the higher end of this ladder, at systems capable of learning to perform *multiple a-priori unknown* tasks, in *multiple a-priori unknown* environments.

Current engineering methodologies assume that code is written by humans. Yet to design systems that exhibit even modest levels of autonomy after they leave the lab, it is clear that we cannot continue on building the whole knowledge base of such systems by hand: The demanding requirements of complex environments, of these future systems' missions, and of their task-execution and problem solving skills, are pushing the development effort out of the reach of what standard software and hardware engineering practices currently are capable of supporting. To build systems of greater complexity than current ones – which any system meeting the above unaddressed requirements unavoidably will be – the methodology must bring the complexity of system development to manageable levels. Ideally we would like a system to gradually implement and improve itself by learning its own task-solving methods using its architecture-provided resources by interacting with its environment, whenever necessary. To move towards systems with higher levels of operational autonomy and increased abilities for autonomous learning, these methodological and systemic shortcomings must be addressed. But if such capabilities are out of the scope of present software development approaches, how shall we approach the engineering of a system endowed with the aforementioned qualities?

In Nivel and Thórisson (2009) we proposed a stringent definition of an autonomous system, as a system that is operationally and semantically closed. Operational closure characterizes a system whose internal agency is maintained – i.e. re-organized and possibly expanded – at runtime by means of its own operation. In the case of software systems, such behavior would mean that components are continually added and deleted as a side-effect of the system‘s normal execution: Some components are learned whereas some others are discounted as a result of their poor performance. From this perspective, such systems can be seen as *autocatalytic* sets[1] whose components are implemented and maintained by the system itself. Semantic closure is a system's ability to control the reorganization of its agency with a purpose: Purposeful, goal-oriented self-organization. To achieve this, a system's architecture must be represented in a semantically transparent manner – making the system *reflective* – to endow the system with an ability to analyze and rewrite its own structure (Nivel & Thórisson 2009, Sanz & Lopez 2000). Autonomy means being free from interference from the outside, which in the case of artificial systems means the system's designers. A system that continuously learns and re-programs itself to adapt and get better must be equipped from the outset with a "seed", an initial and minimal set of instructions general enough to guide the system's bootstrapping its auto-catalytic operation in any instance of its broad set of target environments and tasks. Such a system is *endogenous* in the sense that its behavior results solely from its internal operation and goals and cannot be controlled *directly* by any external operator, but rather can only take *indirect instruction*, which it must evaluate in light of its own current knowledge, like a human would.

The freedom of action entailed by high levels of autonomy is however balanced by hard constraints. First, an autonomous system, to be of any value, is functionally bounded by its mission, which imposes not only the requirements the system has to meet, but also the constraints it has to respect. Second, to keep the system operating within its functional boundaries, one has to ensure that some parts of the system will never be rewritten autonomously – for example, the management of motivations shall be excluded from rewriting as this would possibly allow the transgression of the constraints imposed by the designers. In that sense, the system is also bounded, operationally, by its own architecture. Last, any implemented system is naturally bounded by the resources (CPU, time, memory, inputs) and knowledge at its disposal. For these reasons, autonomy, as we refer to it, shall therefore be understood as *bounded autonomy*.

Our work described here presents evidence and arguments in support of the conclusion that the path towards building artificial general intelligence (AGI) systems can be made possible by meeting – at a minimum – these three requirements of *auto-catalysis*, *endogeny*, and *reflectivity* (AER). These must be implemented in a unified whole, for it is the transversal application of these principles[2] in an architecture that bring their value to the system (Thórisson & Nivel 2009). Unfortunately, none of the methodologies available in the AI or CS literature are directly applicable for designing systems of this nature. For this reason we have advocated what we call a *constructivist AI methodology* (CAIM; Thórisson forthcoming, 2012, 2009, Nivel & Thórisson 2009, Thórisson & Nivel 2009), centered on continuous adaptation – learning that is "always on" and inherent in the system's core operation, and the growth of the system from a small seed. In the main, our constructivist approach has two key objectives: (a) based on the principles of AER, to achieve bounded recursive self-improvement and generality and, (b) to uncover the principles for – and to actually build – systems that, given a small set of seed information, manage the bulk of the bootstrapping work on their own, in environments and on tasks that may be *new* and *unfamiliar*.

---

1 An autocatalytic set is a collection of entities, each of which can be created catalytically by other entities within the set, such that as a whole, the set is able to catalyze its own production. In this way the set *as a whole* is said to be autocatalytic.

In this paper we present a control architecture blueprint – called Autocatalytic Endogenous Reflective Architecture (AERA) – that aspires to enable the engineering of AGI systems. First we describe the assumptions and functional requirements that provided the ground for the design of the architecture. Then in section 3 we describe the principles which guided the design of AERA. Section 4 presents the fundamental principles that govern the execution of AERA-based systems. We provide in section 5 a detailed description of the architecture, both functionally and structurally, and explain how the aforementioned principles are brought to bear to yield higher-order cognitive functions. AERA has been implemented and used for building a prototype the evaluation of which is given in section 6. In section 7, we discuss our approach in light of related work. Finally, in section 8, we sketch out our plans for future developments.

# 2 Assumptions & Requirements

Our objective is to design control architectures for autonomous systems meant ultimately to control machinery (like for example robots, power grids, cars, plants, etc.). All physical systems have limited resources, and the ones we intend to build are no exception: they have limited computing power, limited memory, and limited time to fulfill their mission. All physical systems also have limited knowledge about their environment and the tasks they have to perform for accomplishing their mission. Wang (2011) merged these two assumptions into one, called AIKR – the assumption of insufficient knowledge and resources – which then forms the basis of his working definition of intelligence: “To adapt with insufficient knowledge and limited resources”. We have adopted this definition as one of the anchors of our work, being much in line with Simon’s concept of “bounded rationality” (Simon 1957). This perspective means that we cannot expect any optimal behaviors from our systems since their behaviors will always be constrained by the amount and reliability of knowledge they can accumulate at any particular point in time. In other words we can only expect from these systems their displaying of a best effort strategy.

No system can be built (or build itself) completely from scratch and we assume that a system is given some initial “innate” knowledge that allows interacting with the world – even if in a minimalistic way at first – to acquire more knowledge and improve its performance. Note that we do not consider any assumptions or requirements that pertain specifically to the biological reign. Our work is thus not "biologically inspired" in any interesting or important sense of that term, and it is not our aim to mimic in some way the human mind or biological systems.

[2] The principles are applied system-wide, that is, at *every level of detail*, to *all* the processes that ultimately realize the system’s operation.

We assume that the domain our system operates in is rich, dynamic and open-ended. This means that the domain cannot be fully described in advance, due to its inherent complexity and because it changes continuously, possibly entering states never observed before and for these reasons unbeknown to the system and its designers. That being said, we still assume that the domain presents some regularities of course: The environment is not random and is governed by laws that maintain it in a relative state of stability, at least for long enough to allow the system observing recurrences of states and transitions thereof. However, such regularities can be expected to happen at different time scales, and they are mixed with processes that for all practical purposes can be assumed to be stochastic. The domain is at all times only partially observable: Not all possible states are observable in a given environment at any given moment.

Another assumption is that the kinds of systems we target are not expected to be general problem solvers in the sense of prior aspirations to this end in the world of AI (Newell 1959) and are allowed to rely on ad-hoc specific I/O devices. These devices can be for example, servo-motors, speech recognition, machine vision or inverse kinematics sub-systems, etc. In general, we will consider an I/O device any sub-system for which both its mission and environment (i.e. a sub-set of the global environment the main system operates in) can be well-defined – the rationale being that in these cases ad-hoc optimized solutions are always more efficient than general ones. We do not impose any specific way for integrating the devices into the architecture but the integration should offer various degrees of flexibility. For example, devices can implement full solutions or partial solutions (think for example of many shape recognition sub-systems, federated by the architecture itself; one can also imagine a set of redundant sub-systems with various degrees of performance and reliability). The I/O devices can be considered part of the environment a system has to control and as such the system must be able to model these devices – for example to predict or recognize sensor or actuator failures. This implies that such I/O devices must send to the main system reports of their operation – for example an actuator receiving a command from the system shall answer with the command that was *actually* applied (the desired and actual commands being not always necessarily the same).

The assumptions listed so far form the background for the functional requirements we established for our architecture – which are:

> **R1.** The system must fulfill its mission – the goals and constraints it has been given by its designers – with possibly several different priorities.

Flexibility with respect to top-level goals and constraints is needed if we want the system to be able to focus on the most important tasks at hand, and discard or postpone the achievement of the rest of its primary objectives, when facing scarcity of resources. This also applies to lower-level goals /sub-goals, i.e. goals generated by the system itself to achieve the top-level goals.

> **R2.** The system must be designed to be operational in the long-term, without intervention of its designers after it leaves the lab, as dictated by the temporal scope of its mission.

This means that a system should be as autonomous as possible, to remain operational for extended periods of time without re-adjustment, re-programming, or re-definition from its designers, thus being equipped to face changes in its environment and adapt to these on its own.

> **R3.** The system must be domain- and task-independent – but without a strict requirement for determinism: We limit our architecture to handle only missions for which rigorous determinism is not a requirement.

The requirement of generality also implies that a system must handle inputs coming concurrently from various arbitrary sources and modalities, in other words, no domain-dependent knowledge representation schemes can be tolerated. The systems we envision are essentially non-deterministic in the sense that their behavior depends on the accumulated experience: Given a situation and a set of goals, the machine is likely to perform differently in a new occurrence of said situation from what it did in the past because it may have learned new ways to operate by having captured new knowledge between the two instances of the same situation.

**R4.** The system must be able to model its environment to adapt to changes thereof.

This means in practice that the system must be able to learn new skills, possibly for new environments, augmenting its skill repertoire. The system also has to re-learn some skills it has already acquired in case these are not adapted to the environment anymore. Since changes in the environment are not always predictable, the system must be able to learn continually, incrementally and in real-time. The system does not only have to adapt its behavior but also has to adapt the way it generates it according to the resources at hand. In other terms, the system must be able to adapt its computation to the scarcity of its resources and degrade gracefully accordingly.

**R5.** As with learning, planning must be performed continuously, incrementally and in real-time. Pursuing goals and predicting must be done concurrently.

In particular, the system must anticipate the environment for acting – a controller that does not anticipate its environment is poised to react after the facts, i.e. to lag behind its target. The rationale for concurrent goal pursuit and anticipation is that (a) since achieving goals needs predictions, the latter shall be up to date and thus a system cannot stop predicting because it is planning its next move and, (b) the system will not have the luxury to predict any possible state transition in its environment; the only interesting ones are those that pertain to the achievements of its goals, therefore these goals must be up to date when predictions are generated.

**R6.** The system must be able to control the focus of its attention.

As already mentioned, the system has limited resources and the environment a high level of complexity. A system thus has to dedicate its computing power to only address the stimuli that are most relevant to its goals and discard the rest. For a discussion about the fundamental impact of attention on the control of autonomous systems, see Helgason (2013), Helgason & Thórisson (2012), and Helgason et al. (2013).

From R4, R5 and R6 we can require that the three major high-level cognitive functions - learning, attentional control and planning – must be concurrent.

The system must be able to learn from events whenever they happen, regardless of its other activities; in a similar way, the system cannot wait for learning to terminate before acting, and attention must always and continually be directed appropriately so as to avoid wasting computing power and time, for example by learning irrelevant state transitions and planning according to irrelevant inputs and predictions.

**R7.** The system must be able to model itself.

The system does not only have to model its environment for acting. It should also be able to model itself in the environment to be able to predict its own reactions and the success or failure thereof. Moreover a system also has to model itself for adapting its behaviors to the resources available – for example by changing the priorities of some of its goals and redirect its CPU and time budgets to the most urgent goals. Self-modeling necessitates that part of the operation of the system is visible to the system itself (we call this property operational reflectivity) as internal inputs (the external inputs being the stimuli received from the environment).

> **R8.** The system must be able to handle incompleteness, uncertainty, and inconsistency, both in state space and in time.

Since the environment is only partially observable, and since the system has limited resources, said system can only expect to acquire an incomplete representation of the world, possibly including inconsistencies. Therefore its knowledge shall be defeasible and can only be established to a certain degree and within a certain period of time.

> **R9.** The system must be able to generate abstractions from learned knowledge.

Abstraction is a form of compression and therefore will contribute to maintain the system's operation in its prescribed envelope by reducing the amount of knowledge required to solve some tasks and consequently by reducing the time it takes to process said knowledge. In that respect, the system must be able to learn mathematical functions (a form of abstraction) – for example to predict the next position of a moving object, given its current position and speed. In the work described here we restrict this requirement to linear functions and approximate more complex functions by the recursive application of linear ones on input data.

# 3 Design Principles

In light of our functional requirements outlined above we have established five key principles to guide the design of our architecture.

**Uniform fine-grained executable knowledge.** Knowledge is composed of states (be they past, present, predicted, desired or hypothetical) and of executable code (called models). Models are capable of generating such knowledge (for example, generating predictions, hypotheses or goals) and are executed by a virtual machine (in the case of AERA, its executive).

The granularity of such models shall be kept low for two main reasons. First, it is easier to add and replace small (low-grained) models than larger ones because the impact of their addition or replacement in the architecture will be less than replacement of large models. In other words, low model granularity is aimed at preserving system plasticity, supporting the capability of implementing small, incremental changes in the system. Second, low granularity helps compositionality and reuse; small models can only implement limited low-level functions and, if abstract enough, are more likely to be useful for implementing several higher-level functions than coarser models that implement one or more such high-level functions in one big atomic block. We have referred to this elsewhere as the principle of pee-wee granularity (Thórisson 2012, Nivel & Thórisson 2009, Thórisson & Nivel 2009).

We also need the knowledge to be uniform, that is, encoded using one single scheme regardless of the particular data semantics. This helps to allow execution, planning, and learning algorithms to be both general and commensurate in resource usage (described in more detail below).

The system must be capable of handling vast amounts of knowledge. The system is meant to operate in complex environments where it will be stimulated by a high number of inputs, thus increasing its need to understand them – that is, to produce a large amount of model candidates to predict these inputs. The system is to face novelty and to do so with limited prior knowledge. This means that it is not expected of the system that it is always able to identify relevant inputs; irrelevant or incorrect models shall be expected and the system shall be able to handle these. It follows that such a system has to implement a mechanism to trim down the irrelevant or faulty models, calling for a continuous process of evaluating and revising vast amounts of its acquired but possibly uncertain knowledge.

**Massive fine-grained parallelism.** As we cannot assume guarantees for system down-time (after all, we are targeting high levels of operational autonomy), all activities of the system, from low-level (for example, prediction, sub-goaling) to high-level (like learning and planning), must be performed in real-time, concurrently, and continuously. Moreover, we need these activities to be executed in a way that is flexible enough to allow the system to dynamically (re-)allocate its resources depending on the urgency of the situation it faces at any point in time (with regards to its own goals and constraints), based on the availability of these same resources, over which it may not have complete (or any) control. The approach we chose is to break all activities down into fine-grained elementary reasoning processes that are commensurable both in terms of execution time and scheduling. These reasoning processes are the execution of various kinds of inference programs (models being one), and they represent the bulk of the computing. These programs are expected to be numerous and this calls for an architecture capable of handling massive amounts of parallel tasks (thereafter referred to as reasoning jobs, or jobs for short).

**Experience-based looped-back adaptation and cognitive control.** Working under an assumption of incomplete knowledge and insufficient resources means that the systems we envision are neither likely to have the time and knowledge necessary to accomplish all the jobs they ideally should, given their goals, nor to process every input available in the environment. Standard real-time control systems are designed to be *deterministic*: Capabilities, jobs, and inputs are predetermined, and schedulability - the pre-established proof that *all* jobs will be scheduled for execution - is critical. In sharp contrast, our systems are not expected to be deterministic, as their operation is grounded in their *experience*. Such systems must be able to react at any time, thus possibly striving to *delay* or, even better, *discard* jobs depending on (a) the urgency of the jobs with respect to current or predicted situations, (b) the estimated *value*[3] of the jobs for the system *as a whole* and, (c) the resources available. These estimates (predictions, individual job value and value for the system) are derived from experience accumulated from past operation. From the operational level perspective, a system, as we see it, will never repeat any known algorithm to process a given set of inputs as algorithms are implemented by *learned* models. Due to its continuous learning, the system is essentially always re-computing the way anything should be done, on the fly, thus constantly modifying its "algorithms", even when repeating the same task many times sequentially. From a higher (functional) level perspective however, the system is still expected to behave as deterministically as possible, i.e. to achieve its goals under its prescribed constraints consistently and reliably, adapting to the change of conditions by accumulating experience that will predictably yield more value in the future.

A cornerstone of our approach is that cognitive control results from the *continual value-driven scheduling of reasoning jobs*. According to this view, high-level cognitive processes are grounded directly in the core operation of the machine resulting from two complementary control schemes. The first is top-down: Scheduling allocates resources by estimating the global value of the jobs at hand, and this judgment results directly from the products of cognition – goals and predictions. These are relevant and accurate to various extents, depending on the quality of the knowledge accumulated so far. As the latter improves over time, goals and predictions become more relevant and accurate, thus allowing the system to allocate its resources with a better judgment; the most important goals and the most useful/accurate predictions are considered first, the rest being saved for later processing or even discarded, thus saving resources. In that sense, *cognition controls resource allocation*. The second control scheme is bottom-up: *Resource allocation controls cognition*. Shall resources become scarce (which is pretty much always the case in our targeted system-environment-mission triples), scheduling narrows down the system's attention to the most important goals/predictions the system can handle, trading scope for efficiency and therefore survivability – the system will only pay attention to the most promising (value-wise) inputs and inference possibilities. Reciprocally, shall the resources become more abundant, the system will start considering goals and predictions that are of less immediate value, thus opening up possibilities for learning and improvement. System's resources are poised to continually oscillate between scarcity and abundance, as scarcity will push for more efficiency, which in turn releases more resources that become available for more speculative reasoning jobs, which in turn consumes more resources.

[3] "Value“ is an assessment of "utility“, which refers to the learned, predicted achievability of *all* the goals the system pursues, which is ultimately sanctioned by, and grounded in, the environment. It is *not* the eponym concept, as widely used in the "machine learning" literature, which refers to some intrinsic, axiomatic, hard-coded heuristics. See section 4.3 Equation 6, for an unambiguous definition of the concept in the context of our work.

**Operational reflectivity.** A system must know what it is doing, when, and at what cost. Enforcing the production of explicit traces of the system's operation allows building models of said operation, which is needed for self-control (also called meta-control). In that respect, the functional architecture we seek shall be applicable to itself, i.e. a meta-control system for the system shall be implementable the same way the system is implemented to control itself in a domain. This principle is a prerequisite for integrated cognitive control (Sanz & Hernandez 2012).

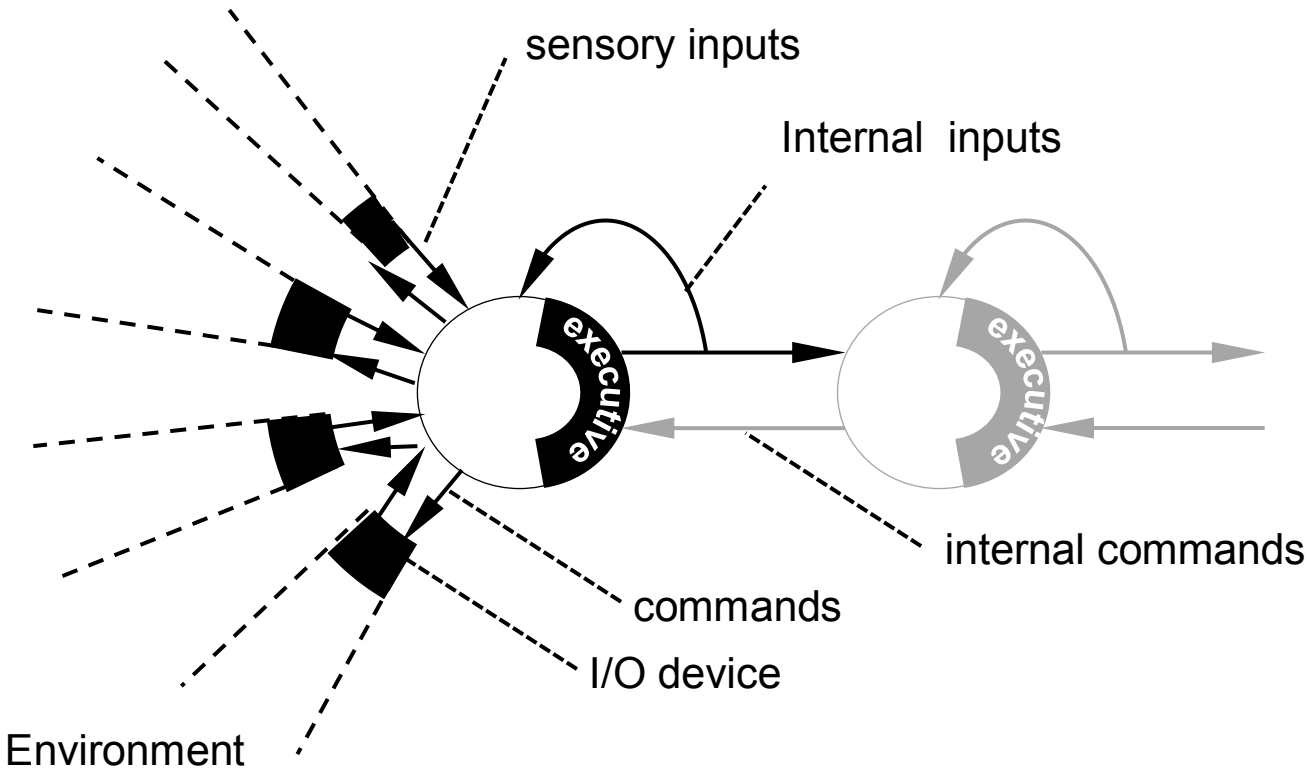

*Figure 2 – Integrated Cognitive Control*

Our approach assumes that the system is interfaced with its environment via I/O devices, dedicated, domain-dependent sensors/effectors. Due to the assumed high ratio of available environmental data to cognitive computing resources, the devices cover only parts of the whole environment at the desired granularity, ranging from coarse-grain (like blob detection from visual data supplied via cameras) to fine-grain (e.g. edge detectors).

The system consists of two essential parts: A memory (white disc) and an executive. The latter executes knowledge in the form of models, based on sensory inputs and the content of the memory (experience).

The executive exposes several parameters and functions that can be viewed as internal effectors and traces of the system's own execution are continuously injected into its memory, in the form of internal inputs. These are processed in exactly the manner as sensory inputs. The processing of internal inputs and of the internal effectors allows – optionally - using an instance of AERA (on the right of the picture) to control another one (on the left), realizing integrated cognitive control (ICC).

**Pervasive flexible representation of time**. Representing time at several temporal scales, from the smallest levels of individual operations (e.g. producing a prediction) to a collective operation (e.g. achieving a mission) is an essential requirement for a system that must (a) perform in the real world and (b) model its own operation with regards to its expenditure of resources (as these include time). Considering time values as intervals allows encoding the variable precisions and accuracies needed to deal with the real world, for example, sensors do not always perform at fixed frame rates and so modeling their operation may be critical to ensure reliable operation of their controllers and models that depend on their input. Also, the precision for goals and predictions may vary considerably depending on both their time horizons and semantics. Last, since acquired knowledge can never be certain, one can assume that "truth" – asserting that a particular fact holds – can only be established for some limited time, and for varying degrees of temporal uncertainty.

In conclusion, these design principles can be unified under the higher-level principle of “holistic design”. According to this principle, robustness shall be seen as the robustness of an entire situated system as a whole, for example, allowing a whole system to carry on most of its relevant operations while facing environmental adversity without breaking down because one of its sub-parts failed. In our view there are no sub-parts, but a dynamic pool of low-level generic processes that altogether implement the necessary functions for achieving a mission. The core idea is (a) to even out the load imposed by said sub-processes among the available resources and competencies and, (b) to fail gracefully when one or more processes fail or fail to be executed due to lack of resources.

In our approach there are no sub-components called “learning” or “planner” and so on. Instead, learning and planning are emergent processes that result from the same set of low-level processes: These are essentially the execution of fine-grained programs and are thus reusable and shared system-wide, collectively implementing functions that span across the entire scope of the system’s operation in its environment. For example, models generate both goals and predictions, some other programs monitor their success or failure and are thus able to reinforce the system’s confidence about their effectiveness. Now, if we picture a skill as a plan, that is a succession of goals, then we see that such a skill is actually implemented by a set of models. From this perspective, learning a skill results from learning models and their sequence of execution, and this results from both the assessment of the performance of said models and the detection of novelty, which in turn produces new models. Both of these activities are examples of the aforementioned generic low-level processes. High-level processes (like planning and learning) influence each other: For example, learning better models and sequences thereof improves planning; reciprocally, having good plans also means that a system will direct its attention to more (goal-)relevant states, and this means in turn that learning is more likely to be focused on changes that impact the system’s mission, possibly increasing its chances of success. These high-level processes are dynamically coupled, as they both result from the execution of the same knowledge – the core of the system, its models.

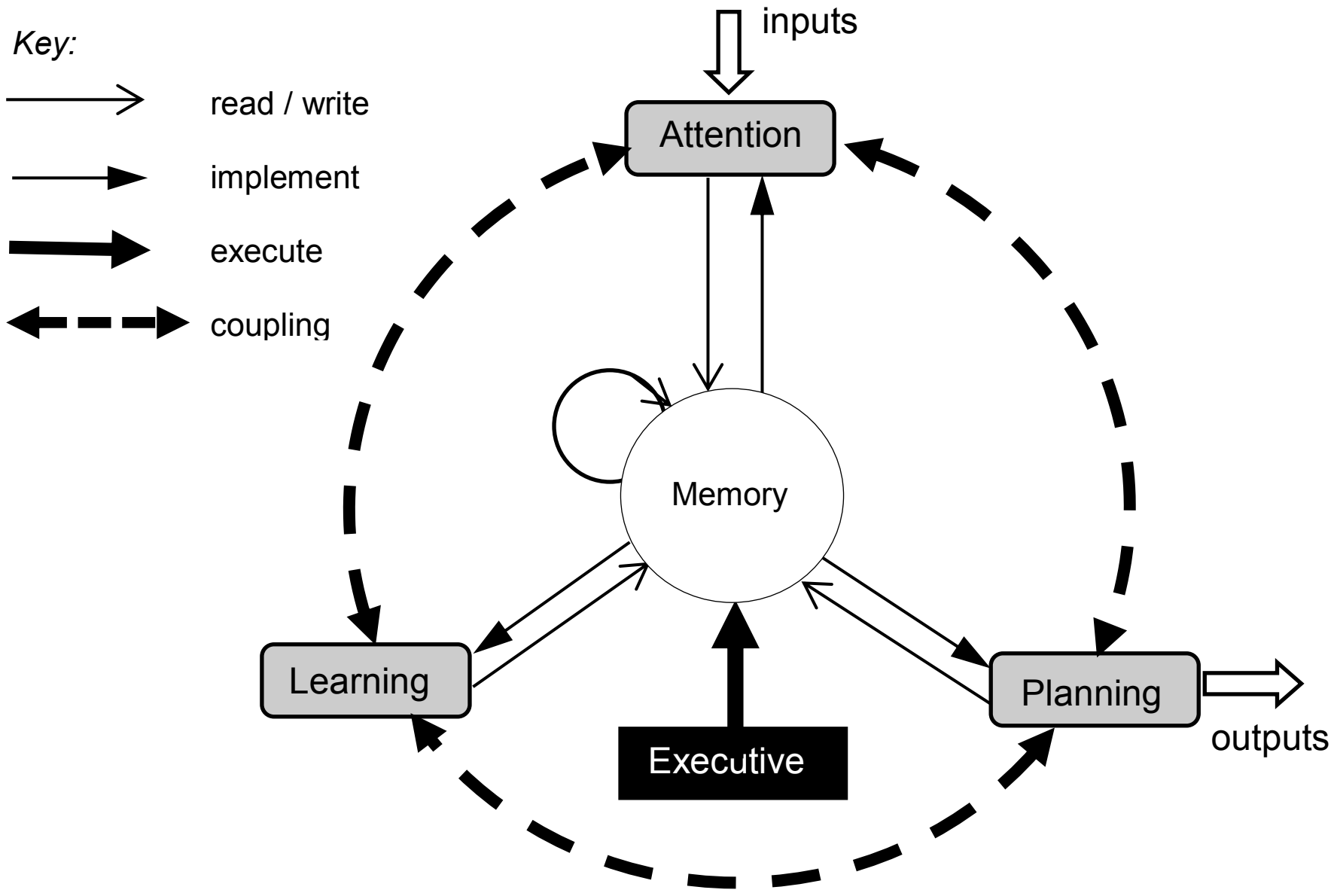

*Figure 3 – Holistic design*

In addition to observed states, assumptions, goals and predictions, the system's memory contains executable code (in the form of models and other programs) and as such, constitutes an active part of the system: It is actually the very core of a model-based and model-driven system. The memory is responsible for most of the computation occurring in the system. Three main cognitive processes – attention, learning and planning – are themselves implemented by programs constituting the adaptable part of the system, the fixed part being the executive. These processes are indirectly and dynamically coupled through the memory as models are added, deleted, activated or phased-out, as dictated by the context and the goals pursued by the system.

# 4 Principles of Operation

We propose four main operational principles for an architecture intended to meet the functional requirements listed in section 2: (a) A way to represent uncertain, defeasible, time-dependent knowledge, (b) a control hierarchy constituted of executable knowledge, either given or learned, (c) a way to deliver real-time performance, anytime, based on the value of computation expenses for the predicted welfare of the system *as a whole* and, (d) a way to abstract knowledge to contribute to the maintenance of computation demands within a limited resource budget. These form the content of the four sub-sections in this section. In the last sub-section, on abstraction, we give a concrete example of operation that pulls together many of the key principles listed here.

## 4.1 Knowledge Representation

Our approach to knowledge representation has its roots in a non-axiomatic term logic. This logic is non-axiomatic in the sense that knowledge is established on the basis of a system's experience, that is, truth is not absolute but rather established *to a certain degree* and *within a certain time interval*. In our approach the simplest term thus encodes an observation, and is called a *fact* (or a *counter-fact* indicating the absence of an observation). A fact carries a payload - the observed event -, a likelihood value in [0, 1] indicating the degree to which the fact has been ascertained and a time interval in microseconds – the period within which the fact is believed to hold (or, in the case of a counter-fact, the period during which the payload has *not* been observed). Facts have a limited life span, corresponding to the upper bound of their time interval. Payloads are terms of various types, some of which are built in the executive, the most important of these being *atomic state*, *composite state*, *prediction*, *goal*, *command*, *model*, *success/failure*, and *performance measurement*. Additionally, any type can be defined by the programmer, and new types can be created by I/O devices at runtime. We will look at each of these in turn:

**Atomic state.** An atomic state encodes a simple relation between objects of the form "entity property value" (for example "entity_12 has_color_component_blue 128", where "has_color_component_blue" is a domain-dependent user-defined term). A counter-evidence of this state could be for example "entity_12 has_color_component_blue 96", both evidences and counter-evidences are expected and AERA has been designed to handle such inconsistencies by allocating its resources based on the priorities of their respective processing (see section 4.3).

**Composite state and instantiated composite state.** A composite state is a compound of abstract facts. Abstract facts are facts where some of their values have been replaced by variables. A composite state is therefore best viewed as a *pattern of the conjunction* of several facts. To take an example, a composite state coded as the conjunction of the facts "E has_color C", "E bears_number N", "E bears L" and "N bears_number L" can describe a bus, with a color, its bus line number and a license plate. When the executive observes an instance of the conjunction of facts specified by a composite state it produces an instance of said state (for instance, "yellow thing1, bearing both line number 19 and thing2 bearing number PDH2O"), as the payload of a new fact. This fact's likelihood is the least likelihood found among its components and its time interval is the intersection of the components' respective time intervals.

**Prediction.** A prediction denotes a hypothetical future state and is encoded as a fact holding a payload (also a fact, the future state). The time interval associated with the predicted future state describes the interval within which the predicted state is expected to hold.

**Goal.** Like a prediction, a goal is a fact holding a payload fact referencing the desired state. The time interval associated with the desired state describes the interval within which the state is to be achieved.

**Command.** A command is an operation which, when embedded in a goal, is to be executed by an I/O device. Issuing a command to a device triggers an answer from said device in the form of the command that has *actually* been executed, which is dependent on the current capabilities of the device (the answer will be less than the command when the latter exceeds the capabilities of the hardware, as expected) – this answer is thereafter called an *efferent copy* and is encoded as the payload of a fact, i.e. as a regular input.

**Model and instantiated model.** A *model* encodes procedural knowledge in the form of a *causal relationship* between two terms. A model is built from two patterns, left-hand and right-hand. When an instance of the left-hand pattern is observed, then a prediction patterned after the right-hand pattern is produced, and reciprocally, when an instance of the right-hand pattern is observed (such an instance being a goal), then a sub-goal patterned after the left-hand pattern is produced. Each time a model predicts, the executive produces a new term, called an *instantiated model* that references the input, the output and the model itself. An instantiated model is thus a trace of the execution of a model and, being the payload of a fact, constitutes an (internal) input for the system. Models form the very core of an AERA system and their operation is detailed in the next sub-section.

**Success/failure.** The assessment of the success or failure of a goal or a prediction (and therefore of the models which produced them) constitutes an (internal) input of the system. These are encoded as fact payloads.

**Performance measurement.** The system periodically assesses its processing performance in terms of the observed lag with regards to meeting its timely targets; for example, achieving goals in time, assessing the performance of its models, assessing the travel time of its jobs in the scheduler (see section 5.3). Such measurements are also internal inputs to the system. As for anything else, performance measurements are encoded as fact payloads.

The life cycles of models and composite states are more complex than those of facts. Essentially their life span and capacity to operate are governed by the entire architecture, based on their individual performance. Models are said to be *reliable* when they predict correctly and consistently so; reliable models tend to be executed more often than unreliable models, the latter being eventually discarded when their performance becomes unacceptable (see section 5.2 for details). In addition, a garbage collector deletes models and composite states that have been the least recently used.

## 4.2 Control Hierarchy

AERA is data-driven, meaning that the execution of code is triggered by matching patterns with inputs. Code refers to models (which constitute executable knowledge), that have either been given (as part of the bootstrap code) or learned by the system. As mentioned above, models are structures composed of two terms – a left-hand term (LT) and a right-hand term (LR) – encoding a causal relationship between the two terms; an instance of LT entails the production of an instance of RT. Both of these terms are patterns, that is, terms containing variables. Models support two modes of execution. The first one, called *forward chaining* operates as follows: When an input term matches a LT, the executive produces an output, a prediction, patterned after the RT. In this mode, inputs can be facts holding any kind of payload, except goals. The other mode is called *backward chaining*: When an input goal matches a RT, an output is produced, a sub-goal, patterned after the LT. Additionally, when an input (other than a goal or a prediction) matches a RT, an *assumption* is produced, patterned after the LT. An assumption is a fact whose likelihood value is computed in a particular way (detailed in section 5.2.4). Notice that multiple instances of both forward and backward chaining can be executed *concurrently* by a given model – i.e. a model can produce several predictions from several different inputs while producing several goals and assumptions, from several other inputs at the same time. In addition to their two patterns LT and RT, models contain two sets of equations, called *guards*. These are equations meant to assign values to variables featured in the output, from the values held by variables in the input. One set of guards supports forward chaining, whereas the other one supports backward chaining. In our current implementation, guards are restricted to linear functions.

The bootstrap code - the initial resource for the system - contains (among other things) drives and top-level models. A drive is an "innate" top-level goal given by the programmer, and whose semantics can also be of a constraint. A drive is essentially a goal whose payload is a fact that cannot be observed – think for example of the drive "keep operating successfully": it is very unlikely that the environment will ever produce explicit evidences of such a state. That is where top-level models come into the picture: these help a fresh AERA-based system get started learning in a new domain. More specifically, a top-level model is hand-crafted for giving the system a way to entail the success (or failure) of a drive from an *observable* (such an observable could be "your owner gives you a reward"). As an AERA-based system is event-driven, drives and top-level models form together the system's motivation, providing a top-down impetus for the system's running, while sensors provide an influx of data, driving its operation bottom-up.

Depending on their respective patterns, models form control hierarchies based on pattern affordances: The output of one model can match a pattern of another and so on (see Figure 4). Such a hierarchy is traversed by two concurrent flows of information, bottom-up (inputs from the environment at the bottom, to the top-level models) and top-down (from the top-level models to commands).

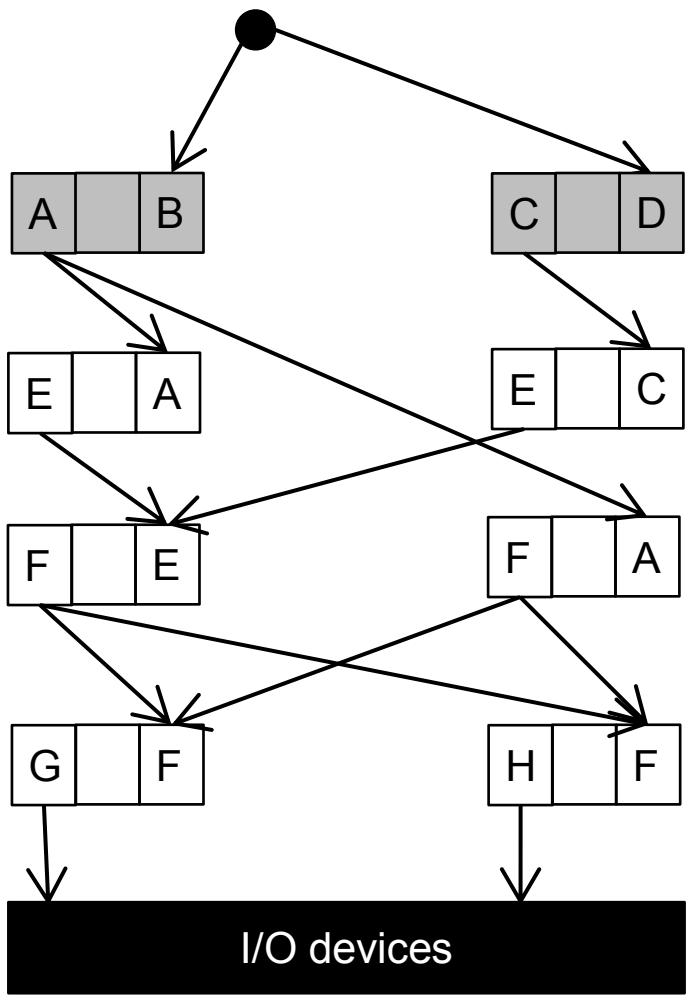

*Figure 4 – Control based on pattern affordances*

A hierarchy of models is depicted. Models are built from two patterns, left-hand and right-hand. These patterns are denoted using capital letters (their parameters and guards are omitted here for clarity). Top-level models are hand-crafted (as part of the bootstrap code), the vast majority of the rest of the models, which for a deployed system can run in the thousands, is learned. Motivated by drives (one is represented here as the black dot at the top), models produce sub-goals when super-goals match their right-hand pattern, and these sub-goals in turn match other models' right-hand pattern until a sub-goal produces a command for execution by I/O devices. In parallel to this top-down flow of data, the hierarchy is traversed by a bottom-up data flow, originating from inputs sensed by the I/O devices that match the left-hand patterns of models, to produce predictions that in turn match other models' left-hand patterns and produce more predictions.

Whenever a model produces a prediction, the executive also produces a corresponding instantiated model: This is a term containing a reference to the model in question, a reference to the input that matched its LT and a reference to the resulting prediction. Such a reflection of operation constitutes a first-class input – i.e. an observable of the system's own operation - which is, as any other input, eligible for abstraction (by replacing values with variables bound together by guards) thus yielding a pattern that can be embedded in a model.

When a model $M_0$ features such an instantiated model $M_1$ as its LT then, in essence, $M_0$ specifies a post-condition on the execution of $M_1$, i.e. $M_0$ predicts an outcome that is entailed by the execution of $M_1$. In case the LT is a counter-evidence of a model's execution (meaning that the model failed to execute *because despite having matched an input, its pre-conditions were not met* – pre-conditions are explained immediately here below), the post-condition is referred to as a *negative post-condition*, *positive* otherwise. Symmetrically, when a model features an instantiated model as its RT, it essentially specifies a pre-condition on the execution of the embedded model instance, i.e. when a condition is matched (LT), the model predicts the success or failure of the execution of a target model (the one an instance of which is the RT). More specifically, what a pre-condition means is "if the target model executes, it will succeed (or fail)". In case the RT is a counter-evidence of a model's successful execution (predicted failure), the pre-condition is referred to as a *negative pre-condition*, *positive* otherwise.

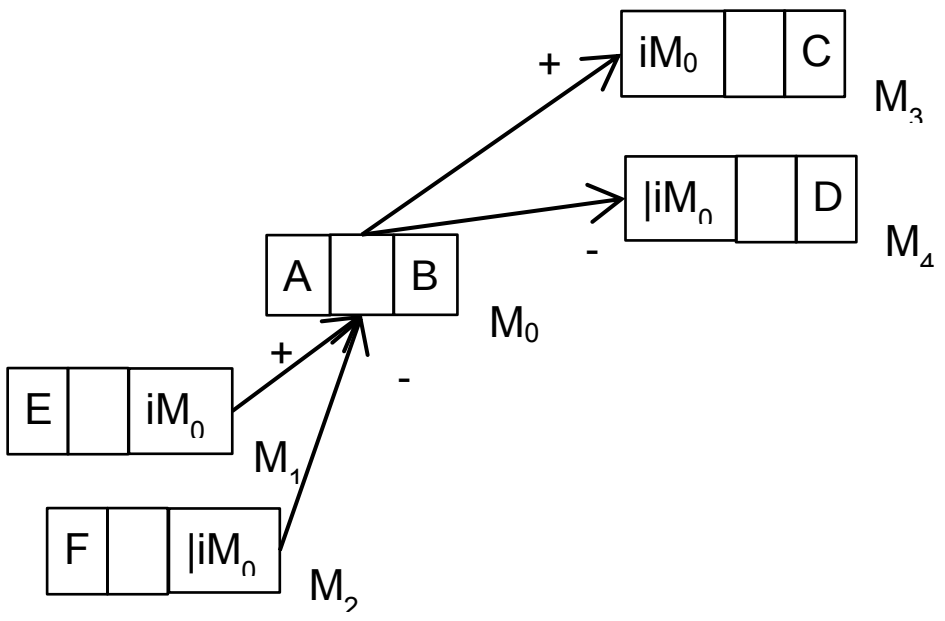

*Figure 5 – Control with pre- and post-conditions*

Each time a model produces a prediction, the executive injects a trace of its execution called an instantiated model (noted $iM_0$ for the model $M_0$). The execution of the positive pre-condition $M_1$ enables the execution of $M_0$ whereas the execution of the negative pre-condition $M_2$ inhibits it.

The execution of $M_0$ is an input matching the left-hand pattern of the post-condition $M_3$, thus triggering a prediction patterned after C, the right-hand pattern of $M_3$. If $M_0$ matches an instance of its left pattern A but its pre-conditions are not met, then the executive produces an input $|iM_0$, which means "failure to execute". This input can match the left pattern of a model (here $M_4$) and trigger some prediction (that would be patterned after D).

Control with pre-conditions consists of ensuring that *all* negative pre-conditions and *at least one* positive one are satisfied before deciding to let the controlled model operate. This decision is made by comparing the greatest likelihood of the negative pre-conditions to the greatest likelihood of the positive ones.

## 4.3 Scheduling

A *job* in AERA is a request for processing one input by one program (for example, a model). All jobs (like for example, forward and backward chaining) are assigned a *priority* that governs the point(s) in time when they may be executed. Jobs are uninterruptible but might get delayed and even eventually discarded if they become irrelevant. Jobs' priorities are continually updated, thus allowing high-value new jobs to get executed before less important jobs, and old jobs to become more valuable than newer ones as new evidences constantly accumulates. Thus a job priority depends on the *utility* value of the program and the expected value of the input (these values are explained below). Value-driven scheduling stands at the very heart of our design and underpins our aim of looped-back adaptation and cognition.

Chaining jobs are some of the numerous jobs AERA can schedule; we will encounter more of these below. Our present concern is to describe the operation of a model hierarchy, which is best understood by looking at the scheduling of the chained jobs.

Inputs are assigned a control value called urgency, defined as follows:

$$THZ(x,t) = \begin{cases} late_deadline(x) - t, & t < late_deadline(x) \\ 0, & otherwise \end{cases}$$

$$Urgency(x,t) = 1 - \frac{THZ(x,t)}{\underset{i}{\mathrm{Max}}\big(THZ(x_i,t)\big) + U}$$

Where $THZ$ stands for "time horizon" and where $t$ is the time of evaluation of the functions, $x$ an input, $late_deadline(x)$ the upper bound of $x$'s time interval and $x_i$ all the inputs in the system. $U$ ($U > 0$) is a parameter of the system meant to prevent the urgency of the input with the highest time horizon from being zero.

*Equation 1*

Inputs, in case their life time expires, are not deleted if they have been scheduled for processing and the corresponding job is still in the jobs list and has not been cancelled. A model is assigned a control value, its reliability, defined as the number of times the model predicted correctly (i.e. the positive evidences of its correct operation) divided by the total number of prediction attempts (total number of evidences) plus one:

$$Reliability(m,t) = \frac{e^{+}(m,t)}{e(m,t) + 1}$$

Where $m$ is a model, $e^{+}(m,t)$ the number of positive evidences for the operation of $m$ and $e(m,t)$ the total number of evidences, both evaluated at time $t$. The reliability is actually the product of the success rate of the model ( $SuccessRate(m,t) = \frac{e^{+}(m,t)}{e(m,t)}$ ) and its experience ($Experience(m,t) = \frac{e(m,t)}{e(m,t)+1}$).

*Equation 2*

The likelihood of a goal (its likelihood to be reached) or of a prediction (its likelihood to come true) is defined as the product of the reliability of the models that were involved in the chaining having produced said goal or prediction:

$$Likelihood(x,t) = \prod_{i} Reliability(m_i(x),t)$$

Where $x$ is a goal or a prediction, $m_i(x)$ are the models forming the chain that produced $x$ from an initial input (a drive if $x$ is a goal, a sensory or internal input if $x$ is a prediction); $t$ is the time at which the function is evaluated. The likelihood of a sensory/internal input is 1 whereas the one of a drive is defined by the programmer.

*Equation 3*

We define the *expected value* of an input (all kinds, except goals) as the product of its urgency and its likelihood:

$$ExpectedValue(x,t) = Urgency(x,t) \times Likelyhood(x,t)$$

Where $x$ is an input and $t$ the time the function is evaluated.

*Equation 4*

The likelihood of a goal, as defined as above, helps the system qualify its experience as it combines the reliability of models, but another valuable source of information is conveyed by the predictions of reaching a desired state. The likelihood of a goal must therefore be redefined to combine both of these two sources of information, the rationale being to lower the expected value of pursuing a goal if the system is predictably more likely to reach the desired state by other means than deriving sub-goals from the goal in question:

$$Let\ P(x,t) = \operatorname*{Max}_i(Likelihood(p_i,t))$$

$$CombinedLikelihood(x,t) = \begin{cases} Likelihood(x,t), & Likelihood(x,t) \geq P(x,t) \\ 1 - P(x,t), & otherwise \end{cases}$$

$$ExpectedValue(x,t) = Urgency(x,t) \times CombinedLikelihood(x,t)$$

Where $x$ is a goal, $p_i$ the predictions of $x$'s target state and $t$ the time the function is evaluated.

*Equation 5*

The priority of a forward chaining job (matching an input with the LT of a model) is the product of the expected value of the input and what we call the *utility* of a model - that is, the (normalized) maximum of the expected values of the goals the model has produced so far and that have still not been achieved. It is worth noting that the utility of a model accounts for the value of executing the model for the system as a *whole*, as it combines the expected values of all the goals currently pursued by the system:

$$Let\ UnnormalizedUtility(m,T,t) = \operatorname*{Max}_i(ExpectedValue(x_i(T,m),t))$$

$$Utility(m,T,t) = \frac{UnnormalizedUtility(m,T,t)}{\operatorname*{Max}_i(UnnormalizedUtility(m_i,T,t))}$$

$$PriorityForwardChaining(x,m,t) = Utility(m,Goals,t) \times ExpectedValue(x,t)$$

Where $m$ is a model, $T$ a class of outputs (either Predictions or Goals), $x_i(T,m)$ the outputs in $T$ produced by $m$, $m_i$ the models in the system, $x$ an input and $t$ the time the function is evaluated.

*Equation 6*

The priority of a backward chaining job is the product of the expected value of the incoming goal and the utility of the model, the latter being the (normalized) maximum expected value of the predictions produced so far by the model. As in the case of forward chaining, the utility of a model has here also a system-wide significance as it corresponds intuitively, in the present case of backward chaining, to the “focus” of a system on data that may fulfill its “desires” (i.e. drives and derived sub-goals); in other words the system’s *attention* is accounted for by the utility values of all the models in the system and the expected values of its current goals (see section 5.1):

$$PriorityBackwardChaining(x,m,t) = Utility(m,Predictions,t) \times ExpectedValue(x,t)$$

Where the parameters and functions are defined as in Equation 6 above.

*Equation 7*

Conceptually, the scheduler’s operation can now be described as follows: It is a list of jobs and a set of worker threads (in the sense of operating system threads) which pick up the highest priority jobs for processing, as a result of which new jobs can possibly be inserted in the list. Job priorities are re-computed frequently, as they depend on time and on the current activity of the system (essentially, the assessment of the model performance, and the set of the current goals). It is worth noting that some jobs may get delayed repeatedly until their priority drops down to insignificant numbers (for example when the urgency of a goal becomes zero, i.e. when its deadline has expired) and eventually get cancelled. This is likely to happen in situations where either the CPU power becomes scarce or the number of jobs exceeds the available computing power – which is the expected fate of any system limited in both knowledge and resources.

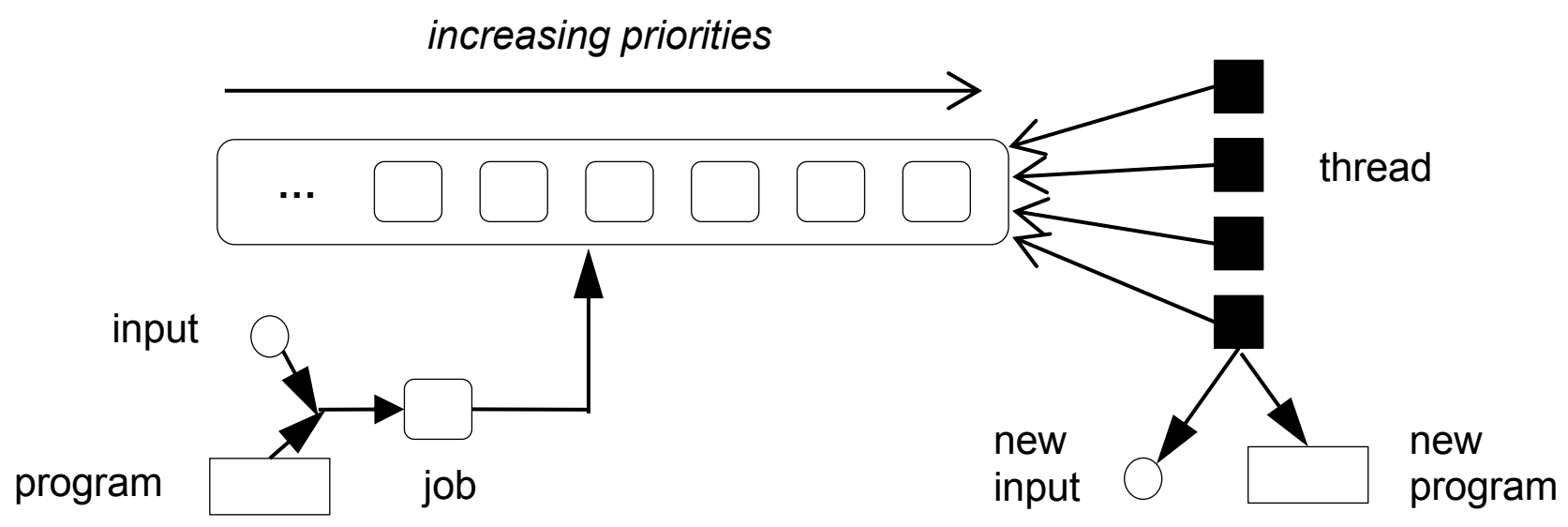

Figure 6 – Scheduling

> The matching[4] of one input and the input pattern of one program triggers the creation of a job which is inserted in the scheduling list. Worker threads extract the highest priority jobs and execute them. This results in the production of new inputs (goals, predictions, reflective inputs, etc.) and programs that fuel in turn the production of new jobs. Notice that priorities are dynamic – they depend both on time and on the ever-changing utility values of all the programs in the entire system - and are thus recomputed frequently.
>
> Some jobs are not controlled by priorities but by time events instead (see section 5.2). The system maintains a separate list of such time-triggered jobs executed by a dedicated pool of threads (not represented here).

In the main, job priorities depend on the past experience of the system (the models and their reliability), the urgency of the inputs and the current activity of the system (in the form of the models' utility values). It follows that, all other values being equal, chaining jobs involving the best models will be scheduled first. Jobs involving inputs that are relevant to the system's operation will also be scheduled first. This is to highlight the fact that the plasticity of computation is achieved by (a) fine-grained jobs and (b) the current activity of the system, which, by design is (c) constrained by (practically unavoidable) limitations on knowledge (number and reliability of the models) and resources (time, memory and available inputs).

In addition to the aforementioned prioritization strategy, we use two ancillary control mechanisms. These come in the form of two thresholds, one on the likelihood of terms, the other on the reliability of models. When a term's likelihood gets under the first threshold, it becomes *ineligible* as a possible input for pattern matching; reciprocally, when the reliability of a model gets under the second threshold, it cannot process *any* input – it is *deactivated* until said second threshold is increased. These thresholds are a filtering mechanism that operates before priorities are computed (the precise operation of these is beyond the scope of the present paper). The executive exposes functions to modify these thresholds as internal commands that can be executed by models, like any other command on effectors in the environment.

[4] Matching is attempted by the executive immediately upon the generation of either a program or an input.

## 4.4 Abstraction

Models are fully abstracted since their patterns contain only variables. This means for example, that any particular input data matching a model's LT will trigger a prediction. A model may thus produce correct predictions for a subset of the matching inputs, and produce incorrect predictions for the rest of these inputs. Now, as we briefly mentioned in section 4.1 above, models are discarded when their predictions become too unreliable. This indicates that over-fitting the inputs can be dangerous for a model: A model that predicts correctly in some cases can still be discarded because it predicts incorrectly in some other cases.

To avoid over-fitting, the scope of models is restricted on a case-by-case basis, meaning that the history of relevant inputs is used to determine whether a model can be executed or not. Relevant inputs are the inputs that have been processed by a model (using forward chaining) and identified as having led to the success or the failure of said model. Case-based control is encoded using models, the only difference with the models as described so far being that they are partially instantiated: Their LT can contain values taken from the historical inputs that have been used to build them. Such partially instantiated models constitute *historical* pre-conditions on the target model (see Figure 7).

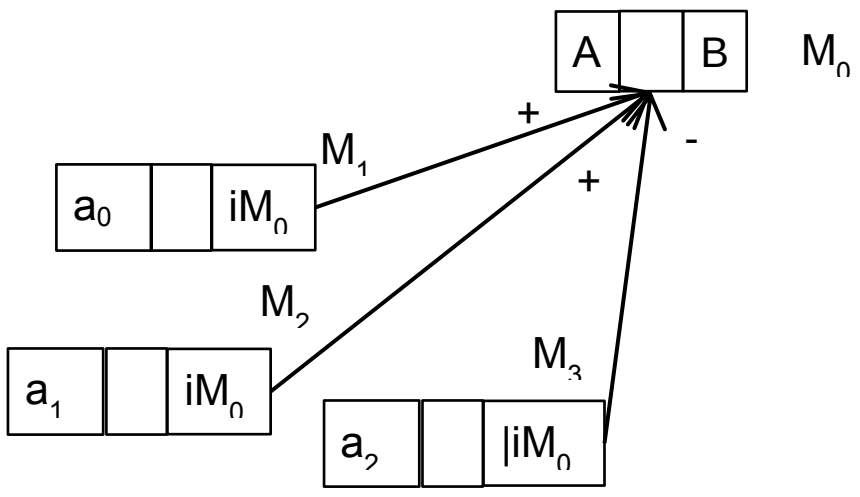

*Figure 7 - Case-based control*

> The history of the execution of the model $M_0$ is captured by other models: These feature left-hand patterns that are less abstracted than the source pattern A in $M_0$: They are the actual inputs that triggered the production of predictions by $M_0$, some having been successful (case of $a_0$ and $a_1$), some not (case of $a_2$ – recall that $|iM_o$ means "failure of $M_0$"). $M_1$, $M_2$ and $M_3$ are historical pre-conditions on $M_0$.

Using partially instantiated models as pre-conditions introduces too much rigidity. For example, in the case depicted in Figure 7, only inputs matching exactly $a_0$ or $a_1$ will allow the model $M_0$ to be executed. This means that any other input differing ever so slightly from the known positive evidences ($a_0$ and $a_1$) will be ignored. If such inputs were later identified as actually leading to the success of the model, then new models would have to be acquired and added to the system, possibly leading to an undesired proliferation of models. We address this issue in the following way. As experience accumulates, the system creates new models representing abstractions of the original partially instantiated models discussed so far. These new models contain more variables than their original and thus are able to match inputs that differ to some extent from known positive evidences; in that sense they are less specified and more flexible than their originals. The construction of flexible models is triggered by the accumulation of new evidences that differ from the known ones while still agreeing on at least one value: Such a new model is copied from the original model and differences between values held by the conflicting evidences are represented by variables introduced in the new model (see Figure 8 for an illustration).

To illustrate the abstraction process, we will take a reasonably realistic but simplified example that explains some of the key principles described so far. Suppose an AERA-based system is interested in knowing which bus to take for going to Reykjavik University. Let $M_0$ be a model saying “any bus of color C, number N and license plate L will go to the university“ (i.e. its left-hand pattern A has three variables C, N and L - its right-hand pattern B being irrelevant for the present discussion). Now assume the system observes four occurrences of the execution of $M_0$, the first two being positive evidences and the last two negative ones: $a_0$ = A(yellow, 19, SX445), $a_1$ = A(yellow, 19, KH203), $a_2$ = A(yellow, 14, PK238) and, $a_3$ = A(yellow, 15, UH714); see Figure 8.

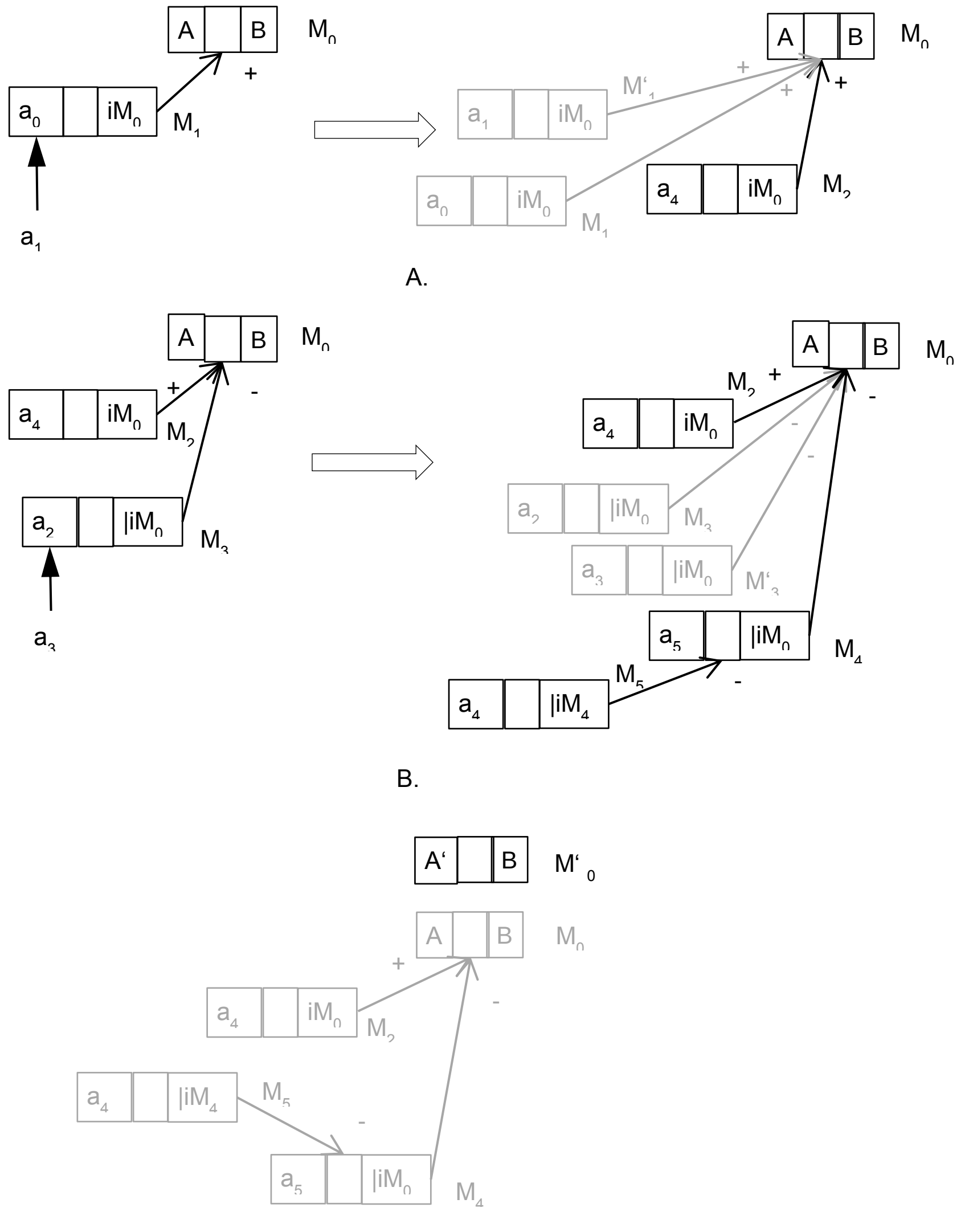

*Figure 8 – Abstraction*

See text for details.

The success of $iM_0(a_0)$ creates the historical pre-condition $M_1$ with $a_0$ as its left pattern. Upon the success of $iM_0(a_1)$, the system attempts to match $a_1$ against known positive evidences, here $a_0$: One value differs (the value of variable L) and a new pre-condition is created: $M_2$ with a left pattern $a_4$ = A(yellow,19,L). The reliability of $M_2$ is computed as the number of times $M_2$ was an effective pre-condition on $M_0$, divided by one plus the number of effective pre-conditions – here 2/3. $M_2$, $M_1$ and $M'_1$ are now three competing pre-conditions on $M_0$ (Figure 8-A) and their fate will depend on subsequent evidences: Shall the particular cases ($M_1$ and $M'_1$) turn out to be actually anecdotal (in our example, the license plate is actually irrelevant) then they will be dismissed on the basis of their relatively poor performance (or usage[5]) and the general case ($M_2$) will prevail – by the means of the standard operation of prediction monitors.

Symmetrically, the failure of $iM_0(a_2)$ creates the pre-condition $M_3$ with the left pattern being $a_2$ (Figure 8-B). Upon the failure of $iM_0(a_3)$, the system attempts to match of $a_3$ against known negative evidences (here $a_2$): Two values differ (the values of N and L) and a new pre-condition is created with $a_5$ = A(yellow, N, L) as its left pattern, $M_4$, coexisting with $M_3$ with a reliability of 2/5, calculated as in case A. The reliability of $M_2$ is recomputed (since new evidences have been observed) and is now also 2/5. Notice that another model, $M_5$, is created and inhibits $M_4$: The negative evidence $a_5$ is an abstraction of a positive one, $a_4$, and therefore $M_2$ shall be prevented to be considered a sub-case of $M_4$. If reality proves that only buses of line 19 actually reach the desired location, then $M_3$ and $M_3$' will eventually be discarded, leaving only $M_4$ to handle counter-evidences.

Flexible historical pre-conditions state, essentially, that some variables are irrelevant. This finds an operational incarnation with respect to backward chaining: Sub-goals targeting atomic states containing at least one variable and derived from pairing super-goals with historical pre-conditions are given an expected value of zero – meaning that they will not trigger further chaining[6]. This translates as follows in our example: Saying the system is "interested" in reaching its desired location (here, an instance of B) means operationally that it pursues a goal G patterned after A. Now, let us assume the pattern A is actually a conjunction of facts – an instance of a composite state whose components (atomic states) would be (a) an entity with a color C, (b) bearing a line number N, (c) with license plate number L. Only backward chaining jobs triggered by the sub-goals targeting (a) and (b) would draw the system's attention – goals targeting (c) would be ignored. $M_0$ would fail to execute, but the goal G would still be achieved unexpectedly, thus triggering the acquisition of a new model $M'_0$: A' → B where A' would be a subset of A, featuring only the variables C and N, but not L as in $M_0$ (Figure 8-C). Section 5.2.2 provides a description of backward chaining through composite states and section 5.2.5 a detailed description of model acquisition. A system operating as in this example with limited resources would naturally focus on the (goal-)relevant inputs and produce more efficient models: (a) $M'_0$ is more focused than the models depicted in case B – it requires less computing power, inputs and time to match its LT (two variables instead of three) and, (b) we would now have only one single model instead of four (which again demands less CPU, inputs, and time).

We close this section on a last technical note about time. One major source of uncertainty is the time at which events occur. Too rigid a temporal prediction production scheme would result in many model failures – the world never repeats itself, at least not at the microsecond scale – and eventually, would result in the dismissal of models that, under less stringent precision constraints, would be more accurate. We address this issue in the following way: When inputs are successfully matched against abstract pre-conditions, the time guards thereof are recalculated using the average of the time intervals of the evidences. If some models eventually turn out to capture generality – meaning in this context that their predictions shall actually be robust to variations of the execution timings – then the aforementioned guard adjustment will trade precision for accuracy. Conversely, if reality calls for precision – i.e. the scope of models is temporally narrower and their predictions more brittle with respect to the execution timings – then this will be reflected in the evidences on which they were based: These will present lesser timings variations and the less general models in question would remain as precise as need be, while retaining accuracy.

---

[5] As introduced in section 4.1, models can be dismissed when they become the least recently used.

[6] Implementation-wise, such goals will not even be produced.

# 5 Architecture

We now turn to describe the set of programs that implement the principal higher-level cognitive functions of the architecture, learning, planning, and attentional control.

As already described, the main components of AERA are its executive and memory, the latter containing inputs (both external and internal), predictions, goals, programs (models, monitors, etc.) and a collection of jobs. The executive uses these elements to (a) maintain and improve the system by adding/removing models to/from the memory, and by (b) controlling the priorities of all the jobs while (c) achieving the goals the system has set itself to fulfill the drives (top-level goals) given by the programmer.

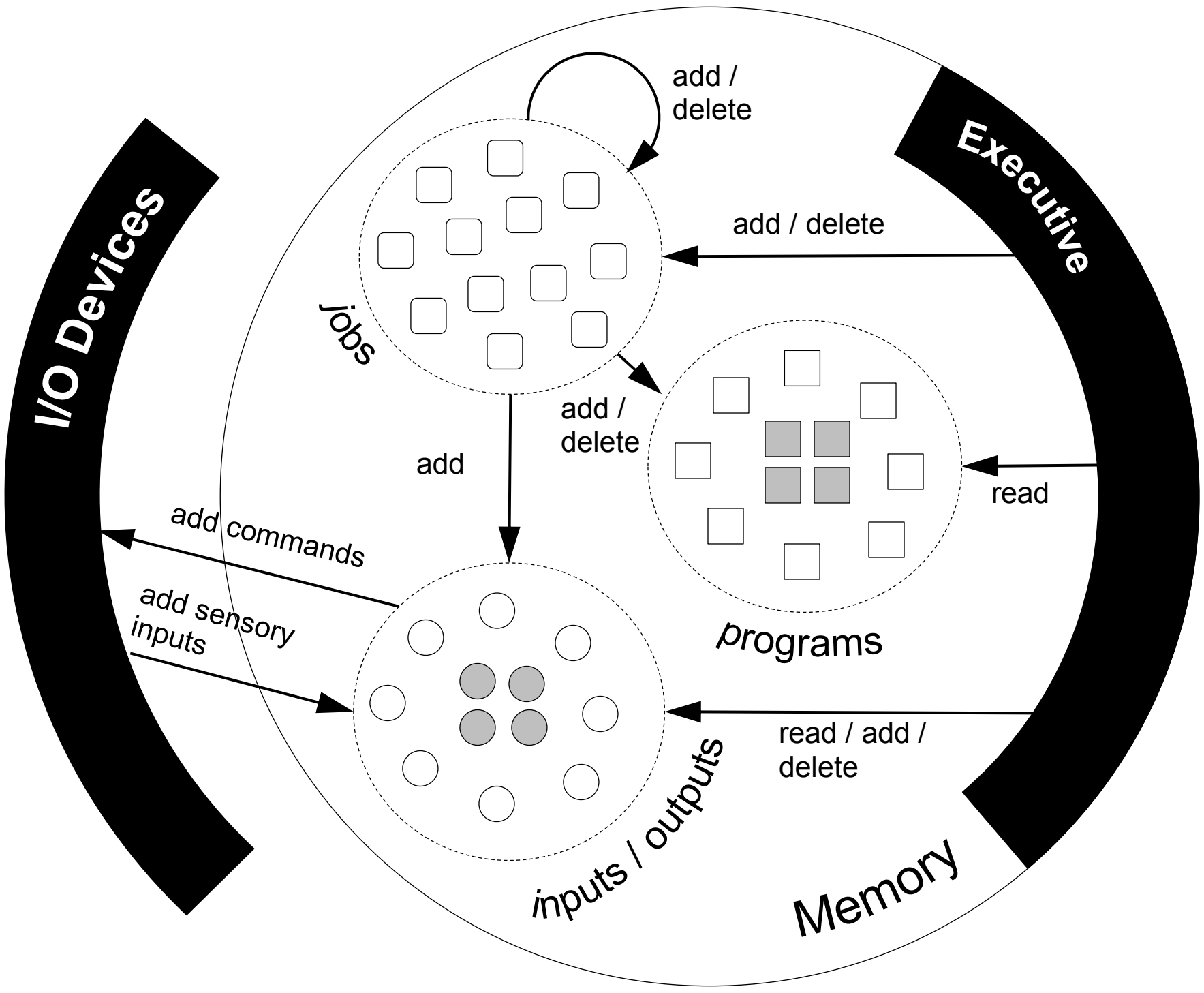

Figure 9 – Architecture

Besides low-level technical components (scheduler, threads, etc.) the executive contains (fixed) algorithms, that are parameterized by jobs, to create programs such as chaining, monitoring, and pattern extraction programs (defined in section 5.2). Models and composite states are also programs, even though they don't result from any parameterization: They are instead either given in the bootstrap code (shown here in grey), or learned. The inputs of the system consist of (a) sensory inputs, (b) goals and predictions produced by jobs and, (c) internal inputs produced by the executive (including instantiated models, instantiated composite states, success/failure of goals and predictions, performance assessments, etc., as explained in the preceding sections). Drives also constitute inputs, and are given in the bootstrap code (in grey).

A running AERA system faces three main challenges: (a) To update and revise its knowledge based on its experience, (b) to cope with its resource limitation while making decisions to satisfy its drives and, (c) to focus its attention on the most important inputs, discarding the rest or saving them for later processing. These three challenges are commonly addressed by, respectively, learning, planning, and controlling the attention of the system. Notice that all of these activities have an associated cost and have to be carried out concurrently. All these activities fit to some extent into the resource- and knowledge budget the system has at its disposal. That is the reason why they have been designed to result from the fine-grained interoperation of a multitude of lower-level jobs, the ordering of which is enforced by a scheduling strategy. This strategy has been designed to get the maximal global value for the system from the available inputs, knowledge, and resources, given (potentially conflicting) necessities. The list of jobs, their purpose, and scheduling priorities, is given in section 5.2.

### 5.1.1 Learning

Learning involves several phases: Acquiring new models, evaluating the performance of existing ones, and controlling the learning activity itself. Acquiring new models is referred to as pattern extraction, and consists of the identification of causal relationships between input pairs: Inputs which exhibit correlation are turned into patterns and used as the LT and RT of a new model. Model acquisition is triggered by either the unpredicted success of a goal or the failure of a prediction. In both cases AERA will consider the unpredicted outcome as the RT of new models and explore buffers of historical inputs to find suitable LTs. Once models have been produced, the system has to monitor their performance (a) to identify and delete unreliable models and, (b) to update the reliability as this control value is essential for scheduling (as described in section 4.3 above). Both these activities – model acquisition and revision – have an associated cost, and the system must allocate its limited resources to the jobs from which it expects the most value. Last but not least, the system is enticed to learn, based on its experience, about its progress in modeling inputs. The system computes and maintains the history of the success rate for classes of goals and predictions, and the priority of jobs dedicated to acquire new models is proportional to the first derivative of this success rate (this is detailed in section 5.2.5).

### 5.1.2 Planning

Planning concerns observing desired inputs (the states specified by goals) by acting on the environment (i.e. issuing commands) to achieve goals in due time in adversarial conditions, like for example the lack of appropriate models, under-performing models, conflicting or redundant goals, and lack of relevant inputs. Planning is initiated and sustained by the regular injection of drives (as defined by the programmer), thus putting the system under constant pressure from both its drives and its inputs. In our approach, sub-goals derived from goals are *simulated*, meaning that as long as time allows, the system will run “what if” scenarios to predict the outcome of the hypothetical success of these simulated goals, checking for conflicts and redundancies, eventually committing to the best goals found so far and discarding other contenders. Here again, goals are rated with respect to their expected value. Simulation and commitment operate concurrently with (and also make direct use of) forward and backward chaining.

### 5.1.3 Attentional Control

A cognitive system generally does not have enough time to process all the inputs it receives, whether from its sensors or in the form of internal inputs, and thus must make a choice to focus its computational resources on the most relevant ones at any point in time. The "interest" or importance of an input depends on the global activity of the system, that is, the set of jobs scheduled to process the input in question: Interesting inputs are the ones processed by high-priority jobs. In our approach, attentional control results directly from its scheduling mechanism, through re-ordering of jobs in accordance with their priorities. Attention, as implemented in in AERA, is transversal and job-*in*dependent: The same attention control is uniformly used for all and any kinds of jobs in the system. For example, attention is required for learning new models but also – at the same time – for planning appropriate courses of action. Attention control is not exclusively based on job priorities, as it results also from selecting proactively sources of inputs in the environment by manipulating the sensing devices. This is achieved by configuring the various sensors the systems has at its disposal by means of issuing commands, computed by backward chaining jobs that are themselves derived from the models, i.e. the procedural knowledge the system has learned (or has been given) so far. All jobs are intertwined in the scheduler's list and therefore contribute altogether to the focus of attention, via the definition of their priority values. As emphasized in section 4.3, job priorities are ultimately derived from the expected value of the system's goals, that is to say that the attentional control is explicitly goal-driven.

## 5.2 Programs

This section presents the various types of programs that implement the cognitive functions mentioned above. As described above, a program is a small component of the architecture that specifies one or several input patterns – for example, a model is one type of program. A job results from the pairing of one input and one program: This is triggered by pattern matching, which is performed continually by the executive, each time an input or a program is injected in the system. Once created, a job awaits execution. If it is eventually executed, the resulting process cannot be interrupted nor preempted at the architecture level.[7] The execution of a job can produce new terms that constitute inputs for the system (for example, goals and predictions), thus triggering pattern matching and therefore the creation of new daughter jobs. Before being executed, a job can be cancelled for various reasons – the most frequent reasons being (a) its input is cancelled, for example, when resolving the conflict between two goals, the system will cancel one of these goals and the jobs for which it was the input, (b) the program it results from is deleted (for example, models can be deleted if their performance becomes unsatisfactory) and, (c) its priority drops down to zero.

Programs are created dynamically by jobs. Creating a program means parameterizing an algorithm with a specific input pattern – such algorithms are fixed and are part of the executive, with the two notable exceptions of models and composite states, which are either defined by the programmer as part of the bootstrap code, or learned. For example, when a job produces a prediction from an input and a model, a monitoring program is created to assess the outcome of that very prediction: It will react to inputs that are evidences (or counter-evidences) of the predicted fact and update the reliability of the model accordingly. Programs are deleted when they become useless; for example, monitors are deleted when the deadline of the expected fact is reached,

---

[7] In our current implementation the thread that a job belongs to is still subjected to the operating system's scheduling events.

and models are deleted when they become too unreliable (see more details on this in section 5.2.3).

The architecture uses two kinds of jobs: priority-controlled jobs (like the ones presented in section 4.3 above) and time-triggered jobs. The execution of the latter is not governed by priorities but is triggered by time events – technically they are locked on timers. These jobs are executed immediately upon triggering and are dedicated to perform timely assessments of expected or desired states. For example, they assert the success or failure of predictions at their deadlines based on the evidences accumulated so far (see sections 5.2.3 and 5.2.4).

### 5.2.1 Model

As already mentioned, a model is a program that, when paired with an input matching its LT, produces a forward chaining job and, when paired with an input matching its RT, produces a backward chaining job. These jobs have been introduced in section 4.3 above and are now complemented with greater details.

Forward chaining through models is the processing of one input matching the LT of a model to produce a prediction patterned after its RT. In addition to the prediction another term is produced, an *instantiated model*, i.e. a trace of the (forward) execution of the model thus instantiated, its input and its output. As mentioned earlier, such a trace of execution constitutes an *internal input* and can be abstracted to form a pattern that in turn can be embedded in any model as its pre- and post-conditions. In case a model has pre-conditions, the executive checks if these conditions are met before executing the model. If so, chaining occurs as described above, otherwise the prediction is produced in *silent mode*, meaning that both the model execution and the prediction itself will be invisible to other programs. The purpose of this mode is to monitor the outcome of the prediction and possibly register an *unexpected success or failure* of the pre-conditions, which will trigger the acquisition of new models (see section 5.2.5). For each prediction, a prediction monitoring program is created (see section 5.2.3) to assess the outcome of the prediction. The job priority for forward chaining has already been given in section 4.3, Equation 6.

Backward chaining through models is the processing of one input matching the RT of a model to produce a sub-goal if the input is a goal or an assumption otherwise. Before committing to a particular goal the system needs to evaluate its possible outcomes to detect and resolve potential conflicts with other goals, and also to select the best next sub-goal when there are multiple ways to achieve it. When a sub-goal is produced from a super-goal it is tagged with a *simulation mark* until it is either cancelled (for example when it conflicts with other more important goals) or committed to. Each time a simulated goal is produced a corresponding prediction is also produced: This prediction is also marked as a simulation and is used by the system to evaluate the consequences of reaching the (simulated) goal in question. Simulated predictions do not trigger the creation of prediction monitors, as no actual event can be expected to confirm the prediction. In case a model has pre-conditions the backward chaining job is not produced before the conditions are met. For each goal a goal monitoring program is created (see section 5.2.4) to assess the outcome of that goal (whether simulated or actual). If the input of the job is not a goal, then the system will *assume* an instance of the LT, with a likelihood computed as for a goal (section 4.3, Equation 7).

### 5.2.2 Composite state

As introduced in section 4.1, a composite state encodes the conjunction of several facts, including facts whose payloads are instances of other composite states, thus allowing the creation of structural hierarchies. A composite state is a program with several input patterns, one per fact. Like models, composite states produce forward and backward chaining jobs when paired with some inputs.

Forward chaining through composite states is triggered by the matching of one input (any kind but goals) with one of the patterns. This in turn produces a new composite state, namely a *partial* composite state, copied from the original, with one input pattern less (the one that has been matched by the input, hence the term "partial") – note that the original is not deleted and remains a legitimate target for matching more inputs. When a partial composite state has only one input pattern left un-matched and matches an input with that pattern, then the executive produces an internal input, e.g. an instantiated composite state, which indicates that an instance of the conjunction of facts specified by the composite state has been observed. If one of the inputs was a prediction then the instantiated composite state becomes a prediction instead of a mere fact. The likelihood of the instantiated composite state is the lowest likelihood of all the inputs that were matched. Composite states are deleted if they have no input pattern left or if the late deadline of one of the inputs they matched has been reached. The priority of a forward chaining job through a composite state is defined as for a model (Equation 6) where the composite state is substituted for the model in the Utility function.

Backward chaining through composite states is triggered in two conditions. First, a goal matching one input pattern of a composite state triggers the production of other goals patterned after the input patterns that have not been matched yet (even though these goals are in effect "side-goals" of the incoming goal, for practical reasons they are treated as sub-goals). The rationale behind this behavior is to entice the system to find situations (conjunctions of facts) that include the goal's target state: Composite states are learned as conjunctions of facts that explain unpredicted states (see section 5.2.5) so it is not a bad idea to try to replicate such a conjunction as targeting one component thereof may entail the observation of another of these components. For example, if we encode in a composite state several attributes of a given entity – say, a bus of a certain color, with a license plate and a line number – then, when trying to find one of these attributes (for example, when trying to spot a bus) uncovering its related attributes may be helpful – in this example, finding a bus will be easier when also trying to find objects with a license plate and a line number. The second case that triggers backward chaining is when a goal targets the instantiation of a composite state instead of one of the state's individual components. In that case, backward chaining occurs as in the first case and applies to the partial states corresponding to the state the system seeks to instantiate. The priority of a backward chaining job through a composite state is defined as for a model (Equation 7) where the composite state is substituted for the model in the Utility function.

### 5.2.3 Prediction Monitor

A prediction monitor has one input pattern, the prediction it monitors. Pairing one actual input (i.e. a sensory or internal input) with the prediction monitor creates a prediction monitoring job. The priority of a prediction monitoring job is the same as the priority of its mother job (a forward chaining job). The purpose of a prediction monitor is threefold.

First it accumulates evidences or counter-evidences of the predicted fact during the predicted time interval, to assess the performance of the model that made the prediction by increasing or decreasing its reliability. Among the accumulated evidences, the evidence (or counter-evidence) holding the highest likelihood value dictates the judgment. If no evidences and no counter-evidences of what was predicted have been observed, then a failure is declared. The assessment is performed by a time-triggered job – a prediction assessment job - at the late deadline of the prediction. This job produces an internal input, a fact indicating the success or failure of the model. In case of a failure, all the predictions that have been produced directly or indirectly from the failed prediction are cancelled and so are the jobs for which they are the input.

If the reliability of a model drops below a threshold $THR_1$ then it is *phased out*: In this mode the model can only create forward chaining jobs and produce silent predictions that will not be eligible inputs to the regular models (i.e. models that are not phased out). Silent predictions are still monitored, thus giving the possibility to improve to a model that was recently getting unreliable. If the reliability of a phased out model gets above $THR_1$, then it is not phased out anymore and resumes its standard operation. When the reliability of a phased out model drops below a second threshold $THR_2$ ($THR_2 < THR_1$) then the model is deleted as are all the programs that were created to manage its productions; the corresponding jobs are cancelled. The way the reliability is defined (see Equation 2) brings some undesired inertia to the switching of modes of operation (phased in or phased out): The more a model has accumulated experience, the more it needs to accumulate future contrary experience to switch modes, because the environment may have changed, making a model temporarily irrelevant but not generally faulty. To address this we implemented a mechanism for reducing switching inertia: In the main, it determines the switch based on reliability computed temporarily using a reference fixed number of evidences[8], instead of the actual and large number of evidences.

The second purpose of a prediction monitor is to maintain abstractions, as described in section 4.4: Upon the success or failure of a prediction produced by a model, the history of the execution of the model is updated, with the possible addition of new pre-conditions.

The third and last purpose of a prediction monitor is to produce evidences of the absence of expected facts: If, at the deadline of the prediction, no evidence of the expected fact has been observed, then the prediction assessment job produces a counter-fact holding the payload of the expected fact. The absence of evidence can result from two causes: Either the environment did not provide such evidences, or it actually did but the system failed to sense them (jobs processing such evidences were cancelled). The likelihood of the latter case is reflected by giving a likelihood value to the counter-fact computed as the utility of the model (see section 4.3, Equation 6).

#### 5.2.4 Goal Monitor

A goal monitor has one input pattern, the goal it monitors. Pairing one input with the goal monitor creates a goal monitoring job. The priority of a goal monitoring job is the same as the one of its mother job (a backward chaining job). Goal monitors come in two variants, one that monitor actual goals (goals already committed to) and one that monitors simulated goals.

The purpose of an actual goal monitor is to check if the goal has been achieved or not. To do so, a goal monitor accumulates evidences or counter-evidences of the desired fact and, at the late deadline of the goal, a time-triggered job – a goal assessment job – is created to establish the outcome of the goal in the form of an internal input indicating the success or failure of the goal. The decision is based on the evidence (or counter-evidence) holding the greatest likelihood value. When a goal succeeds or fails, its sub-goals are cancelled and so are the corresponding chaining and monitoring jobs.

In addition to recording evidences of the success or failure of a goal, a simulated goal monitor also records evidences of the impact of the monitored goal on the success or failure of its super-goal. Such evidences are simulated predictions: Assuming temporarily that a goal's target is achieved, the system predicts the hypothetical outcomes using forward chaining. Simulated predictions reference the simulated goal that triggered them and are eligible inputs for actual goal monitoring as they are required for the actual goal monitor to predict the impact of simulated goals on its own outcome.

[8] Each time the reliability of a model changes, another temporary version of said reliability is computed, using an arbitrary evidence count (defined as a parameter of the system) and a commensurate positive evidence count.

The decision to commit or not to a simulated goal (G) is based on predictions of its impact on actual goals and requires the super-goal ($G_0$) to be an actual goal. The rules of commitment are (using the commit time as the evaluation time):

- If there are no predictions of failure of an actual goal, commit to G.
- If there is a prediction of the failure of an actual goal $G_1$ (i.e. there is a conflict between $G_0$ and $G_1$), and $G_1$ has a greater expected value than $G_0$, then cancel G, otherwise commit to G. Notice that if all sub-goals of $G_0$ were to be cancelled, then $G_0$ would probably fail[9], which would be expected since it would mean that, achieving $G_0$ using the current knowledge conflicts with more important goals.
- If there is a prediction of the success of $G_0$ resulting from a goal $G_2$ that is not a sub-goal of G (G and $G_2$ are redundant) and which has a greater expected value than G, then cancel G.

The decision to commit is triggered by an event described in Figure 10.

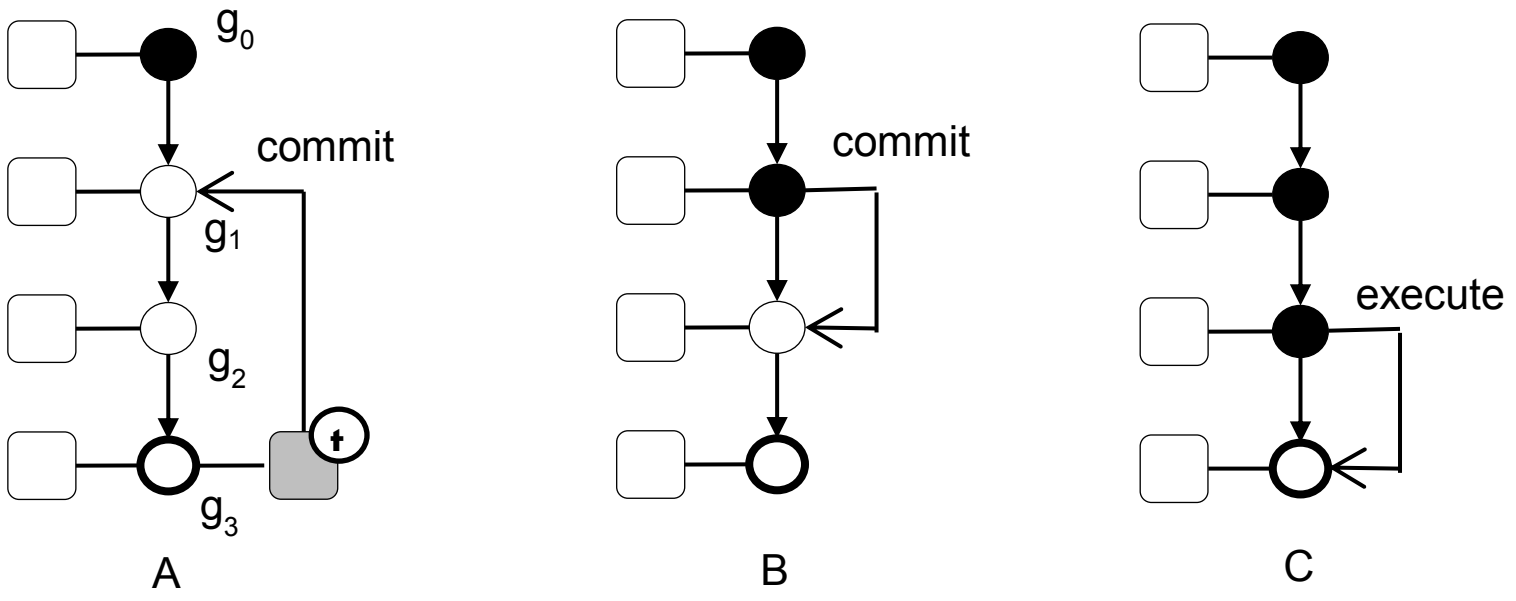

*Figure 10 – Simulation and commitment*

Goals are monitored by jobs (shown as rounded rectangles) accumulating evidences (or predictions thereof) of their target states being reached (for clarity, only one job per goal is shown here). The figure depicts a path in a goal hierarchy constituted by an actual goal ($g_0$) and three successive sub-goals, the last of which ($g_3$) targets a command.

A. If the model having produced $g_3$ has no pre-conditions – or if it does, should these already have been met or predicted – an assessment job (shown in grey) is triggered at the early deadline of $g_3$ and sends a commit signal to the first simulated goal in the path ($g_1$).

B. If the system commits to $g_1$, the commit signal is propagated down the path of the hierarchy $g_3$ belongs to. Otherwise, $g_1$'s goal monitor will attempt to commit upon the matching of relevant inputs, in addition to its standard role of accumulating evidences of the goal's achievement. This additional behavior will cease when $g_3$ is cancelled.

[9] Unless of course an unforeseen event happens to match $G_0$'s target state, in which case $G_0$ would succeed.

C. If the system commit to $g_2$, the command held by $g_3$ is executed. Otherwise $g_2$'s goal monitor behaves as in case B.

### 5.2.5 Targeted Pattern Extractor

A pattern extractor is a program that is generated dynamically upon the creation of a goal or a prediction. Its main activity is to produce models, i.e. explanations for the unpredicted success of a goal or the failure of a prediction: these are called the *signaling events* that trigger the building of models. A single targeted pattern extractor (TPX) is responsible for attempting to explain either the success of one given goal, or the failure of one given prediction. Said goal or prediction is called the TPX's *target*. Under our assumption of insufficient knowledge, explaining in this case is much closer to guessing than to proving, and guesses are based on the general heuristic "time precedence indicates causality". Models thus built by the TPXs are added to the memory and are subjected to evaluation by prediction monitors: Their life cycle is governed essentially by their performance.

A TPX accumulates inputs from the target production time until the deadline of the target, at which time it analyses its buffer to produce models if needed: The TPX activity is thus composed of two phases, (a) buffering relevant inputs and, (b) extracting models from the buffer.

During the first phase, pairing one input with a TPX creates a TPX-accumulation job which merely adds the input to the TPX buffer for further processing. Input relevancy is defined by its utility, computed as the greatest of the utility values of all models in the system that matched the input in question:

$$InputUtility(x,t) = \begin{cases} \operatorname*{Max}_i\big(Utility(m_i(x), Predictions, t)\big), & \text{if positive} \\ U, & \text{otherwise} \end{cases}$$

Where $x$ is an input, $t$ the time the function is evaluated, and $m_i(x,t)$ all the models that matched $x$ with their respective LTs at time $t$. $U$ ($U \in [0,1[$) is a parameter of the system meant to give a chance to inputs that don't match any model to still be considered for model building, if resources allow.

*Equation 8*

The priority of a TPX-accumulation job is proportional to the input's utility, which means that inputs with lower utility will be less likely to be buffered and therefore, that they will be less likely considered for building models. The priority of a TPX-accumulation job also depends on the utility of the model that produced its target, of the incentive of learning said target (defined below) and of a decay function depending on the input found in the buffer. The success rate of the class of goals and predictions it produces are monitored – classes are merely abstracted instances of goals or predictions. The success rate of a class is the number of positive evidences (for goals, the number of times an instance of the goal class was achieved and for predictions, the number of times instances of the prediction class were successful), divided by the total number of evidences (the number of instances of the class). When the target is a prediction, the incentive for learning depends on the unexplained decrease of the success rate of the prediction's class:

$$Let\ SuccessRate(c,t) = \frac{e^{+}(c,t)}{e(c,t)}\ and$$

$$\Delta SR(c,t) = SuccessRate(c,t) - \operatorname*{Max}_{t_r \leq t}(SuccessRate(c,t_r))$$

$$LearningIncentive(c,t) = \begin{cases} -L \times \Delta SR(c,t), & \Delta SR(c,t) < -LTHR \\ 0, & otherwise \end{cases}$$

$$PriorityAccumulation(c, x, t) = LearningIncentive(c, t) \times Utility(m, Goals, t_0(x)) \times InputUtility(x, t)$$

Where $c$ is a class of predictions, $e^+(c, t)$ the number of positive evidences of $c$ being achieved, $e(c, t)$ the total number of evidences both evaluated at $t$, $t$ being the time the function is computed, $x$ a prediction, instance of $c$ and $m$ the model that produced $x$. $t_0(x)$ is the time when $x$ was produced by $m$. $LTHR$ is a threshold defined as a parameter of the system. It is necessary to keep all jobs´ priorities commensurate: $L$ is a parameter of the system defined to set the importance of $|L \times \Delta SR(c, t)|$ relatively to the $Urgency$ function used to compute the priorities of all non-TPX jobs.

*Equation 9*

When the target is a goal, the incentive for learning depends on the unexplained increase of the success rate of the goal's class:

$$LearningIncentive(c, t) = \begin{cases} L \times \Delta SR(c, t), & \Delta SR(c, t) > LTHR \\ 0, & otherwise \end{cases}$$

$$PriorityAccumulation(c, x, t) = LearningIncentive(c, t) \times Utility\big(m, Predictions, t_0(x)\big) \times InputUtility(x, t)$$

Where $c$ is a class of goals, $x$ a goal, instance of $c$, and all other parameters and functions are defined as in Equation 9 above.

*Equation 10*

At the deadline of the target, buffering stops, and the buffer is analyzed as follows, when the target is a goal (the procedure is similar for predictions, see below):

1 If one input is the trace of the execution of one model that predicted the goal's target state, abort – this means that the success was already predicted.

2 Remove any inputs that triggered any model execution.

3 Remove any inputs that were assembled in composite states.

4 Reorder the buffer according to the early deadlines of the inputs[10].

5 For each input remaining in the buffer create a TPX-extraction job, the purpose of which is to assemble a new model from the input and the target.

Shall the target be a prediction, on the other hand, step 1 would be:

1 If one input is the trace of the execution of one model that predicted a counter-evidence of the prediction's target state, abort, as the failure was already predicted.

Reaching step 5 triggers the second phase of TPX activity, where models are built from inputs found in the buffer. The construction of a new model is performed - by a TPX-extraction job - as follows, when the target is a goal:

1 The target is abstracted[11] and forms the RT of a new model (let's call it $M_0$). If the input assigned to the TPX-extraction job is synchronized with other inputs (that is, if their time intervals overlap), then all these inputs are assembled into a single new composite state: this new state is chosen as the LT of the model. Otherwise, the input is abstracted and forms the LT of the model. Notice that new states are identified when their parts are needed for the models being built (instead of resulting from blind temporal correlation): Using composite states as models' LT instead of just atomic states fosters the building of structural hierarchies.

[10] The buffering time of an input is the time at which a TPX-accumulation job was executed, which may not be the same time as the deadline of the input

[11] Here, abstraction means replacing values by variables.

2 If, in a model, some variables in the RT (or in the LT) are not present in the LT (or in the RT), then the job attempts to build guards (see below) to bind these to known variables; otherwise, stop.

3 If some variables in a model are still not bound, then if the buffer is still not exhausted, goto step 4; otherwise, goto step 5.

4 The job considers the next older input to build another model ($M_1$) whose RT is an instance of $M_0$; the unbounded variables in $M_0$ are passed from $M_1$ to $M_0$ as parameters of $M_0$. The execution of $M_1$ allows the execution of $M_0$: M1 is a positive pre-condition on $M_0$. Goto step 2.

5 All models are deleted that hold variables representing deadlines that are unaccounted for, i.e. variables that cannot be computed neither from the LT or RT, nor from the model's parameters list. These models are deleted since they would produce predictions with unbound deadlines, i.e. predictions that cannot be monitored.

If the target was a prediction, then the signaling event would be a failure. Technically this means that $M_1$ mentioned above would have its RT being the failure of the execution of $M_0$, in other words, the execution of $M_1$ would inhibit the execution of $M_0$: $M_1$ would then be a negative pre-condition on $M_0$.

As with TPX-accumulation jobs, the priority of a TPX-extraction job is a function of the utility of the model that produced its target and of the incentive of learning said target. It also depends on a decay function. When the target is a prediction, the TPX-extraction job's priority is:

$$LDecay(t,x) = \begin{cases} \dfrac{t_0(x) - t}{LTHZ} + 1, & t < t_0(x) + LTHZ \\ 0, & otherwise \end{cases}$$

$$\begin{aligned} PriorityExtraction(c,x,t) \\ &= LearningIncentive(c,t) \\ &\times Utility(m, Goals, t_0(x)) \times LDecay(t,x) \end{aligned}$$

Where $c$ is a class of predictions, $e^+(c,t)$ the number of positive evidences of $c$ being achieved, $e(c,t)$ the total number of evidences both evaluated at $t$, $t$ being the time the function is computed, $x$ a prediction, instance of $c$ and $m$ the model that produced $x$. $t_0(x)$ is the time when $x$ was produced by $m$. The $LDecay$ function depends on a system parameter, $LTHZ$, and ensures the job will be cancelled after $t_0(x)$+ $LTHZ$.

*Equation 11*

When the target is a goal, the TPX-extraction job's priority is:

$$\begin{aligned} PriorityExtraction(c,x,t) \\ &= LearningIncentive(c,t) \\ &\times Utility\big(m, Predictions, t_0(x)\big) \times LDecay(t,x) \end{aligned}$$

Where $c$ is a class of goals, $x$ a goal, instance of $c$, and all other parameters and functions are defined as in Equation 11 above.

*Equation 12*

Models contain guards (introduced in section 4.1) and these must be produced from the inputs through induction – for now we limit ourselves to the induction of invertible linear functions of the general form $y = a \times x + b$ where $a$, $x$, and $b$ are either variables of the model or constants. The construction of guards proceeds by attempting to match input data and some guard templates defined by the executive. We will assume that models come in the form: $(LT(Q_0, Q_1, \dots, Q_n, T_0, T_1), RT(P_0, P_1, \dots, P_m, T_2, T_3))$ where $Q_i$ and $P_j$ are variables representing arbitrary quantities, $T_0$ and $T_1$ are variables that define the time interval within which L holds, and $T_2$ and $T_3$ variables defining the time interval for R. The templates currently implemented are given in the following list (an upper case symbol denotes a variable, whereas the same symbol in lower case indicates the value from which the variable has been abstracted; "forward" indicates guards controlling forward chaining; "backward" indicates guards controlling backward chaining).

$Template\ A$

$forward: T_2 = T_0 + period, T_3 = T_1 + period$

$backward: T_0 = T_2 - period, T_1 = T_3 - period$

Where $period = t_2 - t_0$.

$Template\ B$

$forward: P_i = Q_j + Q_k \times period$

$$backward: Q_k = \frac{(P_i - Q_j)}{period}$$

Where $period$ is defined as above.

$Template\ C, only\ if\ LT\ is\ a\ comand\ with\ no\ arguments$

$backward: T_0 = T_2 - period, T_1 = T_3 - period$

Where $period$ is defined as usual (here, the period is actually the duration of the command).

$Template\ D, only\ if\ LT\ is\ a\ comand$

$forward: P_i = Q_j \times Q_k$

$$backward: Q_k = \frac{P_i}{Q_j}$$

Notice that the $Q_i s$ are the command arguments.

$Template\ E, only\ if\ LT\ is\ a\ comand$

$forward: P_i = Q_j + Q_k$

$backward: Q_k = P_i - Q_j$

$Template\ F$

$forward: P_i = Q_j \times c$

$$backward: Q_k = \frac{P_i}{c}$$

Where $c = \frac{p_i}{q_i}$.

$Template\ F$

$forward: P_i = Q_j + c$

$backward: Q_k = P_i - c$

Where $c = p_i - q_i$.

*Equation 13*

Notice that several templates can be used to identify the guards of a single model.

## 5.3 Self-modeling and Self-control

Self-monitoring is the production by the system of data describing the computation of the system. These data are generated automatically by the executive; they are:

- notifications of the productions of models,
- success and failure of goals and predictions,
- notifications of the data used for assembling composite states,
- notifications of instantiated models,
- periodic measurements of the scheduling performance. These are for example the average deadline overshoots for time-triggered jobs, the percentage of jobs cancelled over the total load, along with priority thresholds under which jobs have been cancelled, per kind of job, along with their urgency, the utility value of their program and the likelihood of their inputs.

Such data constitute internal inputs for the system itself and at the elementary level of modeling causal relations, allow capturing pre- and post-conditions on model execution, as introduced in section 4.2. At the higher levels of system control, these internal inputs can also constitute external inputs for meta-systems, i.e. systems in charge of controlling the former one (as introduced in section 3). Periodic measurements of the scheduling performance may indicate possible congestion, i.e. insufficient resources with regards to the jobs at hand. In this case, it is the responsibility of a meta-system (or the system itself) to take contingency measures. These are, for example:

- increase the threshold on terms' likelihood (defined in section 4.3). Goals, predictions and assumptions will be more likely to fall under said threshold, thus limiting the depth of the chaining;
- increase the thresholds $THR_1$ and $THR_2$ that control respectively the switching of phased in/out models and the deletion of the phased out ones (see section 5.2.3). This will decrease the number of models in active duty (phased in) and thus reduce the number of chaining jobs.
- activate the LRU-based garbage collector (introduced in section 4.1). To activate the garbage collector, one specifies a threshold on the least recent time models have been used (e.g. they produced confirmed predictions). Decreasing this threshold reduces the number of models and therefore, reduces the system's load.
- lower the priority of some drives. This will delay or cancel the achievement of the sub-goals stemming from these drives and reduce the overall workload of the system, that is, the number of all the jobs concerned with chaining, monitoring, etc.

Even though self-modeling for meta-control has not been leveraged in our S1 demonstrator, this functionality has been implemented and is operational. As we have argued before (Steunebrink et al. 2013, Thórisson 2012) such functionality is a necessity for a developmental system poised to adapt and make the best use of its resources. With it a system can, for example, model the ways it routinely adopts for achieving some particular goals: This consists of modeling sequences of model execution – these are observable in the form of internal inputs – and, by design, can be modeled using the existing learning mechanisms. The benefit of modeling sequences of execution can be, among others, to enable the system to compile such sequences so as to replace a set of models, which normally have to be interpreted by the executive, with a faster (but also more rigid) equivalent native machine code. Thus *self-compilation* supports a lower-level *re-encoding* of useful and reliable knowledge, which we expect will increase the scalability of AERA.

# 6 Evaluation & Experimental Results

Two experiments are reported here. Experiment 1 (E1) was conceived as a pilot, with the aim of exploring the complexity that the system might be able to handle. In terms of what S1 must learn (described in sections 6.2.1 and 6.3.1), E1 is a strict subset of experiment 2 (E2). In both experiments two humans interact for some time, allowing S1 to observe their behavior and interaction; in both experiments S1's task is to learn how to conduct the interaction in exactly the same way as the humans do, in either role of interviewer or interviewee. The knowledge given to S1 is represented as a small set of primitive commands and categories of sensory data, along with no more than a few top-level goals such as "pleasing the interviewer" (operationally defined as the interviewer saying "thank you" or asking a new question). Details on these are given in section 6.2.

Our results are analyzed using t-patterns (Magnusson 1996, 2000; Jonsson & Thórisson 2010; Thórisson et al. 2013), a method for extracting hierarchies of significant temporal patterns in multi-dimensional source data. Unlike more simplistic statistical measurements, the t-pattern approach we use for analyzing the results of E2 has the unusual feature of giving a rather holistic picture of the overall behavior of the system it is applied to. For readers unfamiliar with this technique we refer to Magnusson (1996, 2000) and Thorisson & Magnusson (2013). The patterns emerging in t-pattern analysis are dependent on three factors: The source material itself, how well it is coded, and the settings of particular analysis parameters. Our source material was a direct recording of the real-time events in the graphical world, i.e. a trace of the human and machine actions in the virtual world frame-by-frame; the coding of the recordings was done by coders with years of experience in this methodology. Parameter settings for all analysis were as follows: Minimum number of occurrences set at 3 and significance level set at 0.0005 unless otherwise noted; other values were set at default (see Magnusson 2006).

First we describe the experimental setup and interaction scenario, then present the results of the two experiments.

## 6.1 Experimental & Scenario Setup

In both experiments S1 observes real-time interaction between two humans in the simulated equivalent of a videoconference: The humans are represented as avatars in a virtual environment (Figure 11) – each human sees the other as an avatar on their screen. Their head and arm movements are tracked with motion-sensing technology, their speech recorded with microphones. Signals from the motion-tracking are used to update the state of their avatars in real-time, so that everything one human does is translated virtually instantly into movements of her graphical avatar on the other's screen. Between the avatars is a desk with objects on it, visible to both participants.

One human is assigned the role of an interviewer (Hq), the other the role of an interviewee (Ha); the goal of their interaction is collaborative dialogue involving the objects in front of them.

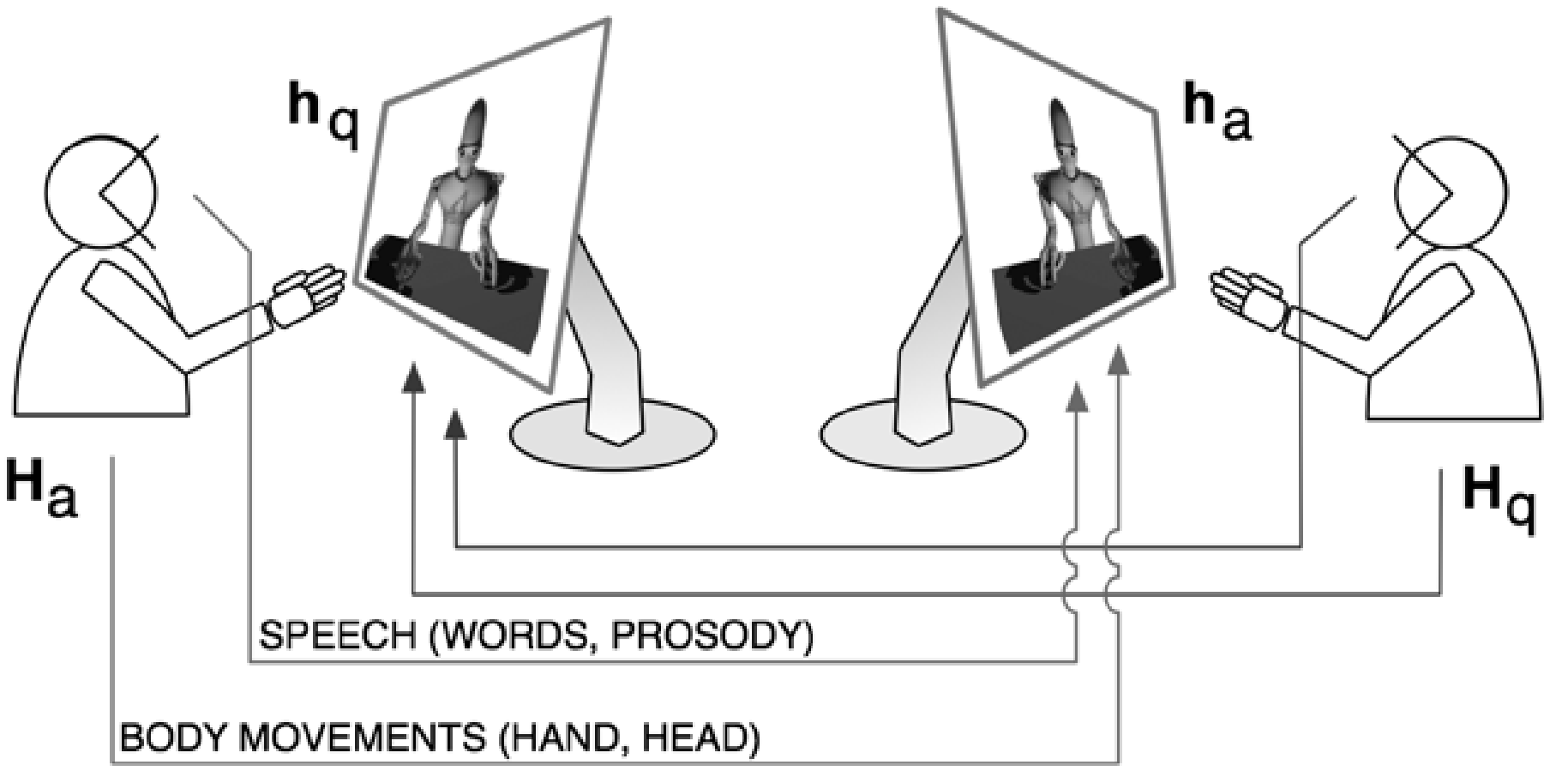

*Figure 11 – Data recording setup for face-to-face interview.*

> During learning by observation S1 observes two humans interacting and subsequently replaces one of them to allow the interaction to proceed as before. The natural interaction between two human participants (H) is mediated via a video-conference virtual-world setup, where participant Hq appears on the monitor of participant Ha as an avatar (hq), and vice versa. One is an interviewer asking questions (Hq), the other an interviewee giving answers (Ha) on the subject of recycling. Their behavior is motion-tracked and used to drive the movements of their avatars; speech is collected via microphones and piped to the speaker on the other's monitor. S1 can replace either human and conduct the interaction in an identical manner.

The data produced during their interaction is represented as follows. Body movements is represented as coordinate changes of labeled body parts of the avatars, in {x, y, z, p, q, r} of the coordinate system of the virtual world used. Each audio signal is piped to two processes: an instance of a speech recognizer (Microsoft SAPI 5.3), and to an instance of the Prosodica prosody analyzer (Nivel & Thórisson 2008). The speech recognition is augmented with timestamps on the words produced (approximate accuracy of time-stamping typically +/-100 ms or better), and (in E2) filtered through a set of 100 allowed words (necessary due to the many false positives produced in live interaction). Words time-stamped with the estimated time-of-utterance are typically output as intermediate hypotheses between 200 and 1000 milliseconds of being uttered, with a "final guess" delivered for each audio segment after a 200 millisecond silence is detected. The prosody analyzer produces time-stamped sound-silence boundaries (accuracy +/- 16 to 32 ms) and prosody information in the form of F0 (with update frequency of 6 Hz; approximate accuracy of 40 ms). This data is the input to S1, streamed to S1 in real-time.

The task assigned to the two humans in E1 is for the interviewer to ask the interviewee to pick up and move things around on the table – a kind of put-that-there scenario (cf. Roy 2005, Bolt 1980). In E1 we provide S1 with top-level goals, contained in the system's "seed" (see section 6.2.1). The seed for S1 in E1 includes five geometrical entities: *cube* and *sphere* of two colors – and a small related set of categories of sensory data, as described in section 6.2.1. In E1 the speech consists of fairly simple utterances which are used to get the interviewee to move the objects around (e.g. "Take the [cube | sphere], and put it here"; "Put [a | the] [cube | sphere] there."); no speech is produced by the interviewee in E1. Also, three gestures were

allowed: grasping/releasing an object, and pointing at a position on the table.

The task assigned to the two humans in E2 is for the interviewer to make the interviewee talk about some properties of objects on the table. The seed for S1 in E2 is also defined as top-level goals, but with more complex natural language, more gestures and more requirements and constraints than in E1.

## 6.2 Experiment 1

### 6.2.1 Experimental Goals & Hypotheses

The goal of E1 was to provide a multi-dimensional task to be learned by S1 through observation, while keeping the complexity of each mode (speech, gesture, semantics) to a defined subset of the complexity targeted in Experiment 2. The data recording setup for E1 was as described in section 6.1 above, with the two humans providing several minutes of interaction to allow S1 to learn how to do it. The objects that the interaction revolves around are: *two blue cubes, one red cube, one red sphere, one blue sphere*. The seed for S1 in E1 is described in detail in Nivel & Thórisson (2013); in short, it consists of a set of primitive commands (*move hand*, *grab*, *release*, *point at*) and a set of dimensions for the input space (*object type*, *color*, *actor's role*, *speech*). The seed also includes initial knowledge that models the consequences of invoking the primitive commands: these models are for example explaining how the position of the system's hand is affected by invoking the command *move hand* and how a hand and an object are linked together after invoking the command *grab*. The natural language used in E1 consisted of a fixed set of sentence fragments (see Table 1).

| **Words** | **Word Order** |
|---|---|
| verbs: put, take<br>nouns: sphere, cube<br>adjectives: blue, red<br>adverb: there<br>determiners: a, the<br>pronoun: it<br>conjunctions: and, ...<br>interjection (ack): thank you | Utterance: (Part1), Part2<br>Part1: take, [a \| the] noun], (conj)<br>Part1: take, [it \| [a \| the] noun], (conj)<br>Part2: put, [it \| [a \| the] [blue \| red] noun], there, ..., thank you |

*Table 1 - Grammar*

The words and word order used in Experiment 1. Note that a sentence starting with "Take it ..." is not allowed as a first sentence in an interaction, as the use of ellipsis must have a prior referent. (Silence of some measurable length is indicated as "..."; parenthesis means optional.)

Human participants in all scenarios observed by S1 are two of the system's developers, who interacted with each other according to the set of targeted behaviors (see Table 2 - What S1 should learn from observation in E1 Table 7 - High-level summary of the target tasks to be learned by S1 in E2). The sequence of actions and the use of multimodal deictics was free-form, and the interaction was real-time. The human participants tried to not make mistakes, but occasional errors were unavoidable as all sessions were live and non-scripted.

In E1 we hypothesized that S1 could learn at least *some* of the behavior patterns needed for conducting the interaction presented in the human-human scenario of E1. S1 was set up to observe the humans for as long as was required for it to predict accurately all major event types observed in the human-human dialogue. After this observation period S1 interacted with a human for a sufficiently long period to produce videos that could be analyzed using the t-pattern approach.

| Category | What is to be Learned | Expected Behavioral Characteristics |
|---|---|---|
| *Interview gross structure* | This involves essentially the system learning a *role*, driven by the top-level goals needed to act in either one, to the point where it can assume either the role of the interviewer or interviewee without problems. | The interaction (E1) / interview (E2) is a sequence of requests (by interviewer) and responses to those (by interviewee). This pattern should emerge quite clearly. |
| *Turn-taking* | As the system is not intended to learn (complex) language, we look at turn-taking that can be generated, measured, and evaluated without looking at the content (words) of the dialogue. | Efficient turns: Small speech overlaps and short or no silences between turns. |
| *Explicit manual deictics* | Successful resolution of a manual deictic (pointing gesture) by the interviewer allows interviewee to place objects in the right location, and to pick out a referenced object out of the five. | Pointing to objects and places on the table disambiguates speech by providing spatial information. |
| *Ellipsis* | The use of pronoun "it" and "the X" (e.g. "Take the cube" in the beginning of a new instruction) is used to reference an object mentioned earlier. | Correct combination of dialogue events to allows correct uses of pronoun and adverb, supporting disambiguation/indication of what should be done. |

*Table 2 - What S1 should learn from observation in E1*

### 6.2.2 Results of Experiment 1 (E1)

In E1 S1 only needed to observe the humans for a few minutes until its performance was error-free in both roles. Videos of S1 in both roles were then recorded, of comparable duration, for subsequent t-pattern analysis. Results show that the performance of S1 matches the human-human scenario very closely. S1 is able to *proceed as the humans in this interaction scenario, without making mistakes, in either role*. The main high-level results of experiment 1 are summarized in Table 3. At the lower level of implementation, the level of models, S1 learned that state transitions (like moving a hand) can be achieved by asking the interviewee to perform actions instead of performing these itself, as a new addition to its prior knowledge encoded in its seed. S1 also learned that, to enact the transition of an object's state from one location to another, one way is to grab said object first, then apply the same state transition to the hand that holds it – this was achieved by modeling pre- and post-conditions on the model describing the state transition. S1 also learned the sequences of orders ("take a blue cube" then "put it there") to fulfill the imposed drive; it also learned to satisfy this drive twice in a row with different targets (a blue cube first, then a red sphere), as demonstrated by the human actors – this results from the hierarchization of control via model affordances. S1 identified the correlation between deictics and utterances (e.g. "there" correlated with pointing gestures) – this is an example of learned structural hierarchy (in the form of composite states) -, as well as ellipsis ("put *it* there"). The pronoun "it" has been learned to identify the object that draws the most attention (in terms of job priority), i.e. the target of the most valuable goals (picking an object is a learned pre-condition on the next step, moving it to some location, to earn the reward) – this is an example of value-driven resource allocation steering cognition and vice-versa.

| Category | What Has Been Learned | Result |
|---|---|---|
| *Interview gross structure* | S1 has successfully learned both roles of interviewer and interviewee from observation, using only the top-level goals needed to act in either one. S1 has learned how to structure dialogue in an interview, as observed in the human-human interaction. | S1 can conduct dialogue with a human efficiently and effectively, in a way that is virtually identical to human-human interaction. |
| *Turn-taking* | S1 has learned the basic skills of turn-taking form observation, as evidenced by the clearly demarcated turn-taking patterns. | S1 efficiently and effectively takes turns, asking questions at the right times (as interviewer) and answering at the right times (as interviewee). |
| *Explicit manual deictics* | S1 has learned to use some deictics effectively by observation, indicating a referent by pointing when saying "there". | Both the timing and form of the gestures is appropriate for the context. Resolution of a manual deictic (pointing gesture) by the interviewer allows interviewee to place objects in the right location, and to pick out a referenced object out of the five. |
| *Ellipsis* | S1 has learned to use ellipsis in sentence interpretation and generation. | The use of pronoun "it" and "the X" (e.g. "Take the cube" in the beginning of a new instruction) is used to reference (as interviewer) / interpreted (as interviewee) correctly as an object mentioned earlier. |
| *Understand / instruct correctly* | As interviewee S1 is able to pair the correct actions with the multimodal speech and gesture acts generated by the interviewer. | Appropriate and correct actions taken, given the behavior of interviewer / interviewee. |

*Table 3 - Summary of results obtained in Experiment 1 (E1)*

The conclusions are based on videos of the interaction between S1 and a human, in both roles, and on the statistical t-pattern analysis presented in this section.

In E1 the high-level results of t-pattern analysis of S1's behavior can be summarized as follows:

- A high number of complex turn-taking temporal patterns were detected in the individual data sets analyzed, comparable between all conditions (human-human, agent as interviewer, agent as interviewee).

- The number, frequency, and complexity of detected patterns indicate that behavior was highly synchronized in all situations.
- This synchrony was found to exist on all levels of temporality, with highly complex time structures that extended over considerable time spans, where some of the patterns repeated in a cyclical fashion.

When looking at simple and short turn-taking patterns, similar structures are detected across all dyads in the S1 dialogue sessions (S1 as interviewer + S1 as interviewee). More complex patterns are detected in the human-human dyads than in the S1 interaction, as S1's induction process produced abstracted models of the human behavior that were simpler than the observed human behavior, yet achieved the same goals (getting the interlocutor to behave appropriately in the context of the interaction, based on its stated goals).

A thorough inspection of the videos of S1 revealed no errors in the interaction on behalf of S1; the system had acquired and generalized the observed human-human behaviors to a sufficient level to allow it to perform error-free communication in subsequent real-time interaction with a human, under the same operating constraints as in the human-human scenario.

| Observation Time | # Turns Observed | # Errors After Observation Period |
|---|---|---|
| ~2.5 mins | ~17 | 0 |

*Table 4 – Basic statistics in E1*

In E1 S1 only has to observe the human interaction for about 2.5 minutes before it is able to predict accurately their behavior, and subsequently assume either role without making any errors.

| Detected Pattern | Description | Significance |
|---|---|---|
| h2_drop, h1_ack | Interviewee drops an object on the table, interviewer acknowledges with a "thank you". | $p < 0.005$ |
| h1_instr, h1_point | Interviewer gives an instruction and indicates a location deictically (by pointing with hand/finger). | $p < 0.005$ |
| h1_pickup, h1_drop, h2_ack | Interviewee grasps an object and releases it, interviewer acknowledges. | $p < 0.05$ |
| h1_instr, h1_point, h2_pickup, h2_drop, h2_partner | Interviewer gives interviewee an instruction and points out a location deictically, interviewee grasps an object and releases it and looks at interviewer (head facing interviewer's location). | $p < 0.05$ |
| h1_instr, h1_point, h2_pickup, h2_drop, h1_ack, h1_instr, h2_pickup, h2_drop, h1_ack | Interviewer gives an instruction and indicates a location deictically, interviewee grasps an object and releases it, interviewer acknowledges and gives another instruction, interviewee grasps an object and releases it, interviewer acknowledges. | $p < 0.05$ |

*Table 5 - Frequent patterns in all conditions*

Some temporal behavior patterns seen in the data analyzed from Experiment 1 (t-pattern analysis). These relatively frequent patterns were statistically significant ($p<0.0005$), consisting of events (see Table 6 for legend) that occurred frequently in the same sequence with significantly similar timings in all conditions (human-human, S1-as-interviewer and human-interviewing-S1).

| | | | |
|---|---|---|---|
| Hq | human, interviewer role | Ha | human, interviewee role |
| S1q | S1, interviewer role | S1a | S1, interviewee role |
| h_table | head oriented towards table | h_partner | head oriented towards other participant |
| partner | trunk turned towards other participant | point | hand points at object |
| pickup | hand grasps object | put | hand releases object |
| ask | a question is asked | reply | an answer is produced |
| feedb | back-channel feedback is produced | inf | a type of information utterance or action is produced |
| instr | participant gives instruction to other | ack | a reply from other is acknowledged (by uttering "thank you" or by a head nod) |

*Table 6 – Legend*

In the interaction the only back-channel feedback produced was a head nod by the interviewer indicating that the interviewee should continue answering.

In our previous results from an analysis of human-agent interaction with the Gandalf agent (Jonsson & Thórisson 2010, Thórisson 1996, 1999) the patterns seemed to be somewhat slower on average than in natural human-human interaction, which can likely be explained by a longer average duration between critical elements of the turn-taking system, especially the speech recognition. Compared to Gandalf, the pace/tempo of S1's dialogue sessions is much closer to the speed of human-human dyadic interaction, especially in dyadic interaction between subjects who know each other well (e.g. friends). A closer comparison between S1 and other human-human data further suggests that the turn-taking reaches a “mean level” patterning quickly, as is the case in dyadic interaction between friends (Jonsson 2006).

These positive results allowed us to step up the complexity of the task significantly, including more complex speech and interaction, as explained in the next section.

## 6.3 Experiment 2 (E2)

### 6.3.1 Experimental Goals & Hypotheses

Given the success of E1, the goal of E2 was to provide S1 with an appropriate increase in complexity of the task. We defined a scenario that included all behavior included in E1, but with considerably increased complexity in both spatial and language behavior. The input dimensions were encoded as before, with more objects:

*aluminum can, glass bottle, plastic bottle, cardboard box, newspaper and painted wooden cube*. The vocabulary was 100 words with substantially more variety in sentence structures. In E2 S1 was given no grammar. The task of the participants in E2 is to talk about these objects on the desk in front of them, in particular, the interviewer's task is to ask the interviewee about the materials which the various objects are made of, and the pros, cons, cost, and methods for recycling them. As before, one participant acts as interviewer, the other as interviewee.

The data flow setup for E1 was as described in section 6.2.1 above, as is the seed (see also Nivel & Thórisson (2013)). In addition to models included in the seed for E1, the seed for E2 consist of models rewarding the interviewer for having the interviewee speak, and models rewarding the interviewee for speaking.

As in E1, in E2 human participants in all scenarios were not trained actors. They interacted according to the targeted set of behaviors (see Table 7 and Table 8). The sequence of their actions and the use of multimodal deictics was free-form and real-time, the interaction thus semi-improvised. The human participants tried to not make mistakes, but occasional errors were unavoidable as all sessions were live and non-scripted. No formal grammar definition was created for the speech in E2, and neither this, nor a list of permissible sentences, were provided to S1 by the designers before S1's observation sessions started. Due to the number of commission errors in the speech recognizer its output was restricted to the set of 100 permissible words (see Table 8).

We hypothesized that S1 could learn the behavioral patterns learned in E1, and in addition would learn a sufficient amount of new behaviors that were introduced in E2, to the extent that it could conduct at least some of the more complex interaction behaviors required in E2, in either role of interviewer or interviewee.

We had S1 observe the humans until it accurately predicted all major event types observed in the dialogue. We then had S1 interact with the humans for a sufficiently long period to produce videos that could be analyzed with the t-pattern approach; recordings of S1 interacting in either role with one of the humans (same as who participated in the human-human scenario) were tens of minutes each. This formed the basis for subsequent data analysis.

| **Dialog Pattern** | **Details** | **Expected Behavioral Characteristics** |
|---|---|---|
| *Interview: interleaved questions and answers* | In E1 S1 had successfully learned how speech and gesture is used together by the interviewer to get the interviewee to perform multimodal acts. Now, in E2, the interview includes new gestures and speech for both roles, resulting in a more complex interview structure. | S1 should learn how to interleave actions and speech utterances to achieve the goals of the dialogue. |
| *Spatio-temporal reference through manipulation* | Simple manipulation is often used as a tool for disambiguating language. Here the co-incidence of speech, e.g. use of "it" or "this" and a manipulation gesture (picking up / holding an object) can be used to identify the intended reference. | Synchronization of speech and object manipulation such that the latter disambiguates the former. |
| *Language: Sentence construction* | Sentence structure requires an appropriate order of words. Table 8 shows the sentence structures used by humans in E2. | Without the use of grammar, S1 should be able to construct sentences in either role of interviewer and interviewee, based on that |
| | | observed being in the human-human interaction. |
| *Language: Constructing proper answer to questions* | When the interviewer asks a question, not only must the gestures and speech be interpreted for the correct response, the reply constructed must be appropriate to the question. | Given the numerous valid questions that can be asked in E2, S1 should reply with an appropriate and correct utterance. |

*Table 7 - High-level summary of the target tasks to be learned by S1 in E2*

In E2 S1 should learn everything it was seen to learn in E1 (see Table 2), but in E2 we add the additional hypotheses listed in this table.

| |
|---|
| Which releases more greenhouse gasses when produced, [an aluminum can or a glass bottle \| an aluminum can or a plastic bottle \| a plastic bottle or a glass bottle]? |
| What [else \| more] can you [tell, tell me, tell us, say] about [this \| that \| it]? |
| There are many types of plastic. |
| Tell [me \| us] about this [object \| thing \| one] |
| More energy is needed to recycle a plastic bottle than a can of aluminum. |
| Compared to recycled plastic, new plastic releases fifty percent more greenhouse gasses. |
| More energy is needed to recycle a glass bottle than a can of aluminum. |
| A glass bottle takes one million years to disintegrate completely in the sun. |
| Glass is made by melting together several minerals. |
| A recycled aluminum-can pollutes (only) five percent of what a new [can \| one] pollutes. |
| Recycling an aluminum-can costs only five percent of a new one. |
| Compared to recycling, making new paper produces thirty-five percent more water pollution. |
| This is a cube made from unpainted wood. |
| Untreated wood is biodegradable. |
| Plastic is made from petroleum. |

*Table 8 - Example sentence structures used by the human participants in E2*

The vocabulary in E2 contained 100 words (words in parenthesis can be omitted), which were combined in numerous ways by the human participants in E2. No formal grammar definition existed or was created for the speech in E2, and neither this, nor a list of permissible sentences, were provided to S1 by the designers before S1's observation sessions started.

### 6.3.2 Results of Experiment 2 (E2)

In E2 S1 learned everything that it observed in the human-human interactions which is necessary to conduct a similarly accurate and effective interaction.[12] The socio-communicative repertoire acquired autonomously by S1 after an observation period of approximately 20 hours, has been *correctly learned, with no mistakes in its subsequent application, including timing of all actions* (Table 10). This repertoire, including skills in either role (interviewer and interviewee), consists of:

- Correct sentence construction - correct word order.
- Effective and appropriate manual and head deictics (gesturing towards object being talked about at the right time, gazing towards it using head direction when mentioned or pointed at).
- Appropriate response generation; answer (as interviewee) and sequence of questions (as interviewer).
- Proper multimodal coordination in both interpretation and production, at multiple timescales (interview, utterance, and sub-utterance levels).
- Turn-taking skills (avoiding overlaps, avoiding long pauses), and utterance production – presentation of content (answer/question) at appropriate times with regard to the other's behavior.
- Interview skills - doing the interview from first question to last question.

| Category | What Has Been Learned | Result |
|---|---|---|
| *Interview gross structure* | As evidenced in the videos, the interview proceeds exactly as in the human-human condition; questions are sequenced appropriately by S1 and several are posed before the interview is over. S1 also learned to use interruption to keep the interview within the allowed time limits. | S1 has acquired the socio-communicative dialogue skills to conduct a real-time multimodal interview with a human in real-time, and can correctly follow its structure as demonstrated in the human-human dialogue scenario. |
| *Interview: Interleaved questions and answers* | In E1 S1 learned how speech and actions are used together by the interviewer and interviewee, respectively. In E2 the interview includes gestures and speech for both roles. | S1 has acquired the ability to correctly interleave questions and answers in light of the goals of the interview, in either role of interviewer and interviewee. |
| *Turn-taking* | While the turn-taking seems slightly slower-paced than typical human-human interaction, the style and action repertoire that S1 exhibits is precisely that observed in the human-human condition. | S1 has learned basic human turn-taking skills, including as is appropriate for efficient language use. |
| *Explicit manual deictics* | S1 clearly shows proper use of deictics, both using palm as a pointing device and the index finger. Head direction is also used as a deictic device in some cases. | S1 has learned to use deictics. Both the timing and form of the gestures is appropriate for the context. |
| *Spatio-temporal reference* | S1 can pick up an object when talking about it, and refer to it in language with | S1 has learned the meaning of manipulation as a deictic device. The timing and use of |

12 Videos of the interaction can be found on www.humanobs.org and on youtube.com under CADIA's video channel CADIAvideos.

| | | |
|---|---|---|
| *through manipulation* | the pronoun "this". | manipulation for reference is appropriate for the context. |
| *Language: sentence construction* | S1 constructs all sentences correctly – the sequence of words is produced using generalized models acquired autonomously from observing the human interaction. | S1 can construct sentences in either role of interviewer and interviewee, based on those observed in the human-human interaction. |
| *Language: Constructing proper answer to questions* | When the interviewer asked a question, not only were the gestures and speech interpreted for the correct response, the reply constructed was appropriate to the question. | Given the numerous valid questions that can be asked in E2, S1 should reply with an appropriate and correct utterance. |

*Table 9 - Summary of the key dialogue skills learned by S1 in E2*

In E2 the observation period was 20 hours, which is dramatically higher than that required in E1 for an equivalent error-free performance. This is due to the complexity of the sentences the system had to learn in E2, which alone have dramatically greater combinatorics than all of the behaviors observed in E1.

| Observation time | # turns observed | # errors after observation period |
|---|---|---|
| ~20 hrs | ~8000 | 0 |

*Table 10 - Basic statistics in E2*

As in E1, visual inspection of the resulting videos from E2 revealed no errors on behalf of S1 in either role; the system has generalized the human-human scenarios it observed to a sufficient abstraction level so as to make no mistakes when interacting with a human.

The t-pattern analysis bore out the conclusion that S1 had acquired all relevant behavioral skills – in fact, if it wasn't for the synthesized speech it is difficult to identify a difference in performance by S1 and the humans. The plots in Figure 13, Figure 14 Figure 16 show the most frequent patterns seen in E2, asking and replying, in all conditions.

The results of the t-pattern analysis mirror those obtained in E1, showing again a high number of complex turn-taking temporal patterns in the individual data sets analyzed; the number, frequency, and complexity of detected patterns indicate that behavior was highly synchronized in all situations, and synchrony was found to exist on all levels of temporality, with highly complex time structures extending over considerable time spans, and a number of the patterns occurring in a cyclical fashion, as would be expected in a structured interview.

The most frequent pattern is, as in E1, the ask-reply sequence, which contains similar timings across all conditions. In Figures 15, 17 and 18, more complex patterns can be seen. Simple and short turn-taking patterns display similar structures across all three conditions. Somewhat more complex patterns are seen in the human-human condition than those involving S1, due to the somewhat smaller repertoire of behavior displayed by the agent than the humans.

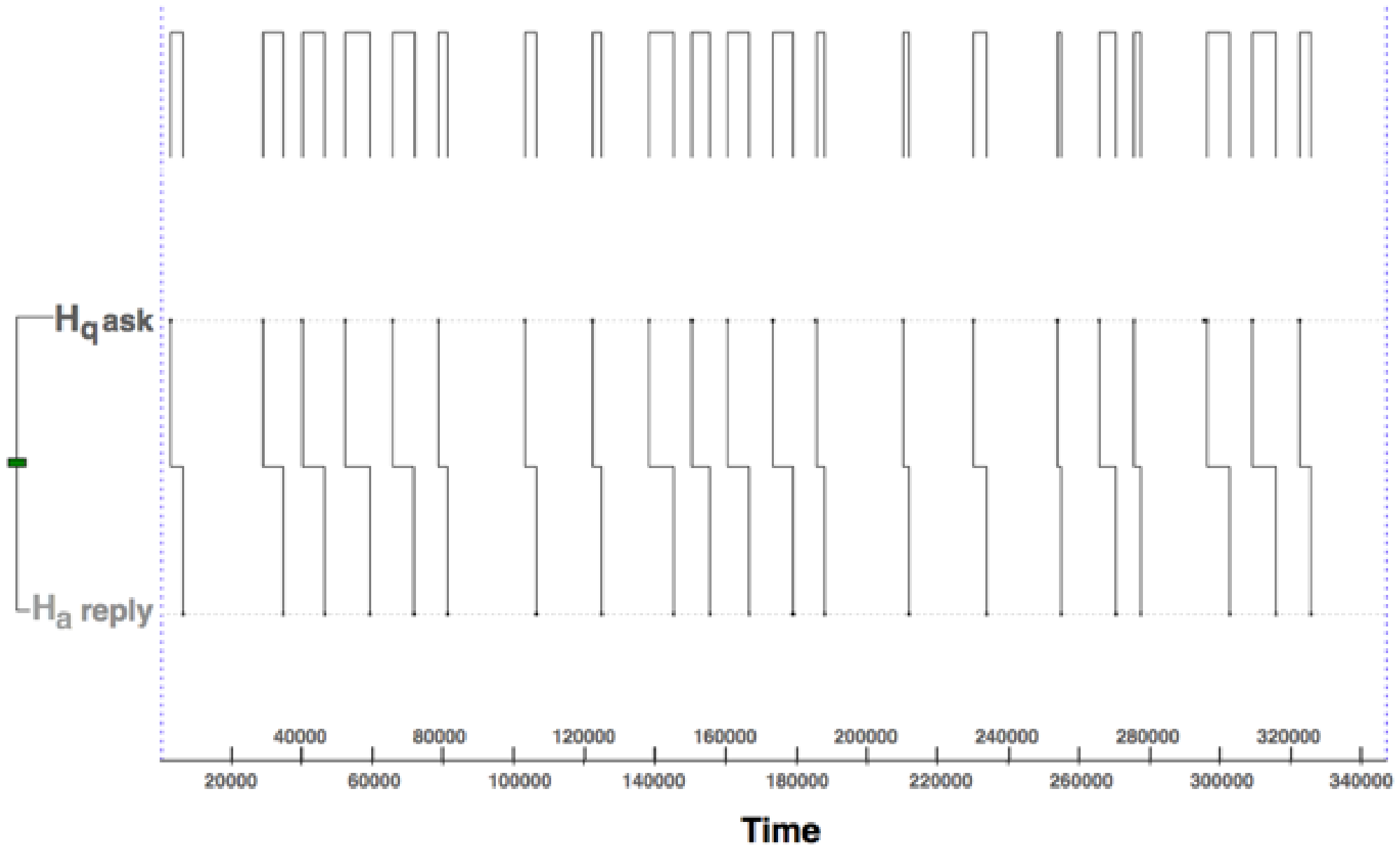

*Figure 12 - Most frequent pattern in human-human condition in E2*

Not surprisingly, the most frequent pattern occurring in the human-human condition is the question-answer sequence. (Timescale is in frames @30fps = 3.1 minutes total. For legend see Table 6.)

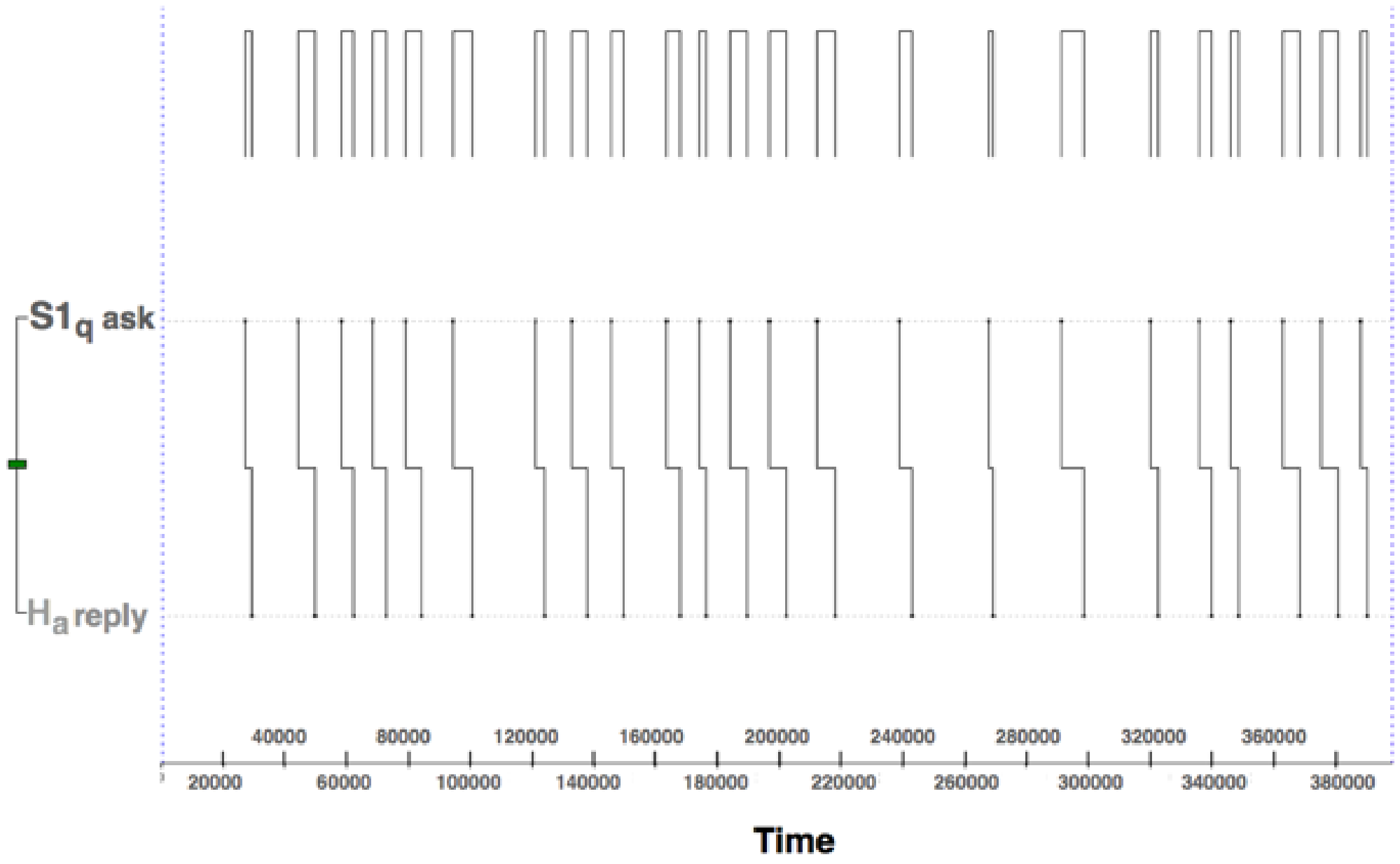

*Figure 13 - Most frequent pattern in S1-as-interviewer condition in E2*

Just like the human-human condition, the most frequent pattern occurring in the S1-as-interviewer condition is the question-answer sequence, in a very comparable manner. (Timescale is in frames @30fps = 3.5 minutes. For legend see Table 6.)

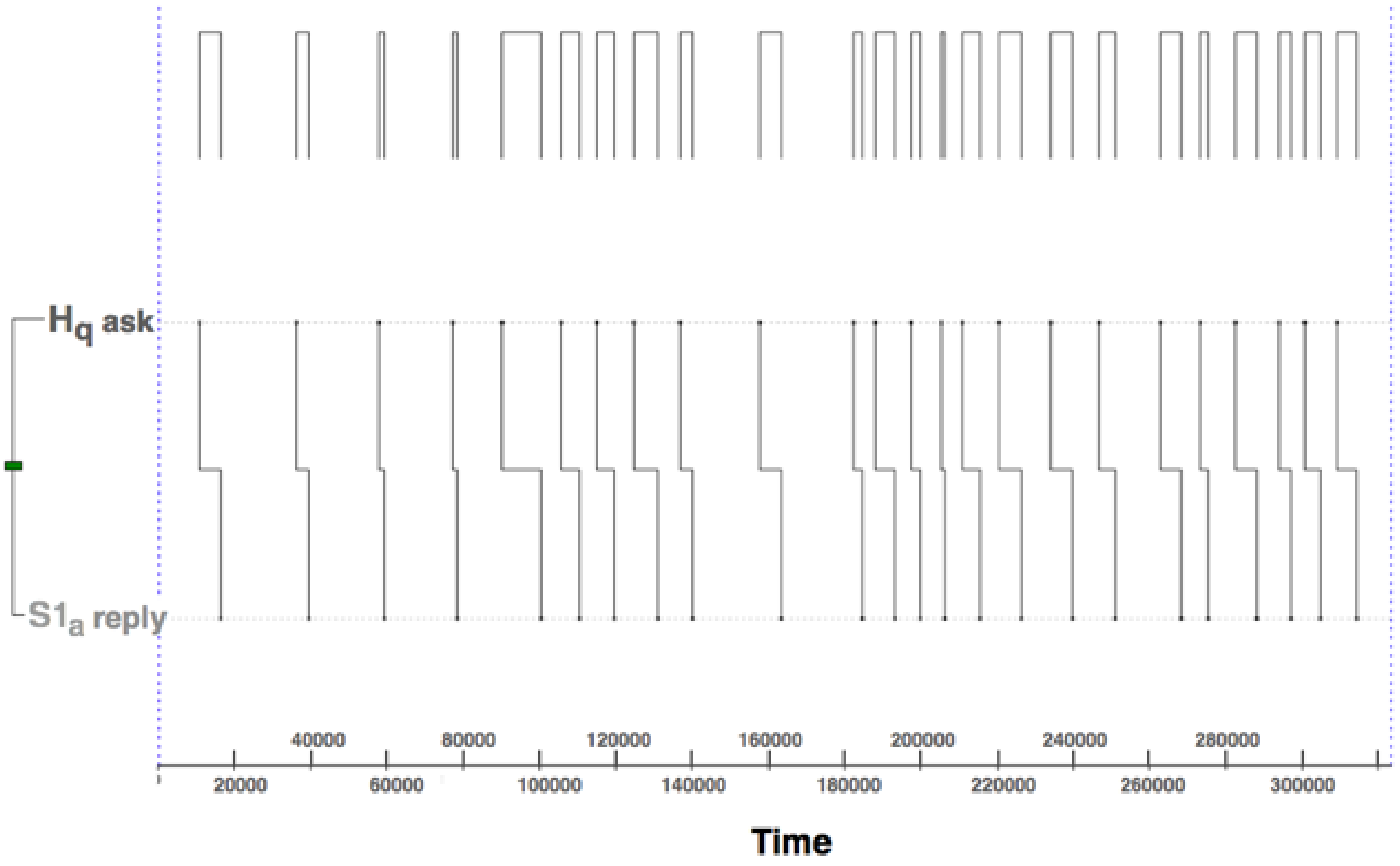

*Figure 14 - Most frequent pattern in human-interviews-S1 condition in E2*

Just like the human-human scenario, the most frequent pattern occurring in the human-interviews-S1 scenario is the question-answer sequence, in a very comparable manner. (Timescale is in frames @30fps = 2.9 minutes. For legend see Table 6.)

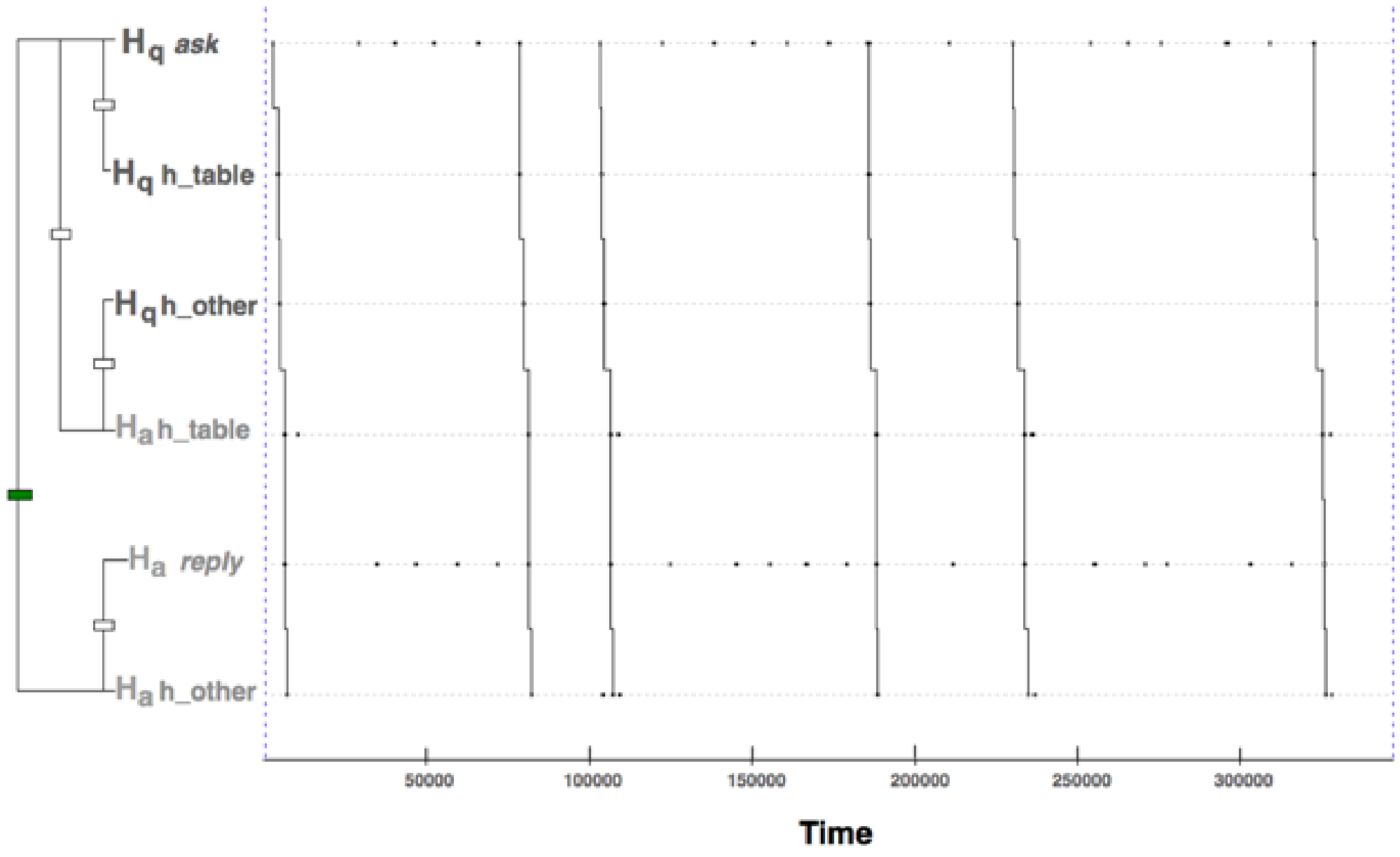

*Figure 15 - Example of common patterns seen in human-human condition in E2, involving question-answering, head direction, and hand activity*

First the interviewer asks a question, looks at the table, then looks back at interviewee, after which the interviewee looks at the table and begins to answer, then looking back at the interviewer. In text format, the tree is the following: ((( h2_ask, h2 h_table )( h2 h_partner, h1 h_table ))( h1 reply, h1 h_partner )). (Timescale is in frames @30fps = 1 minute duration. For legend see Table 6.)

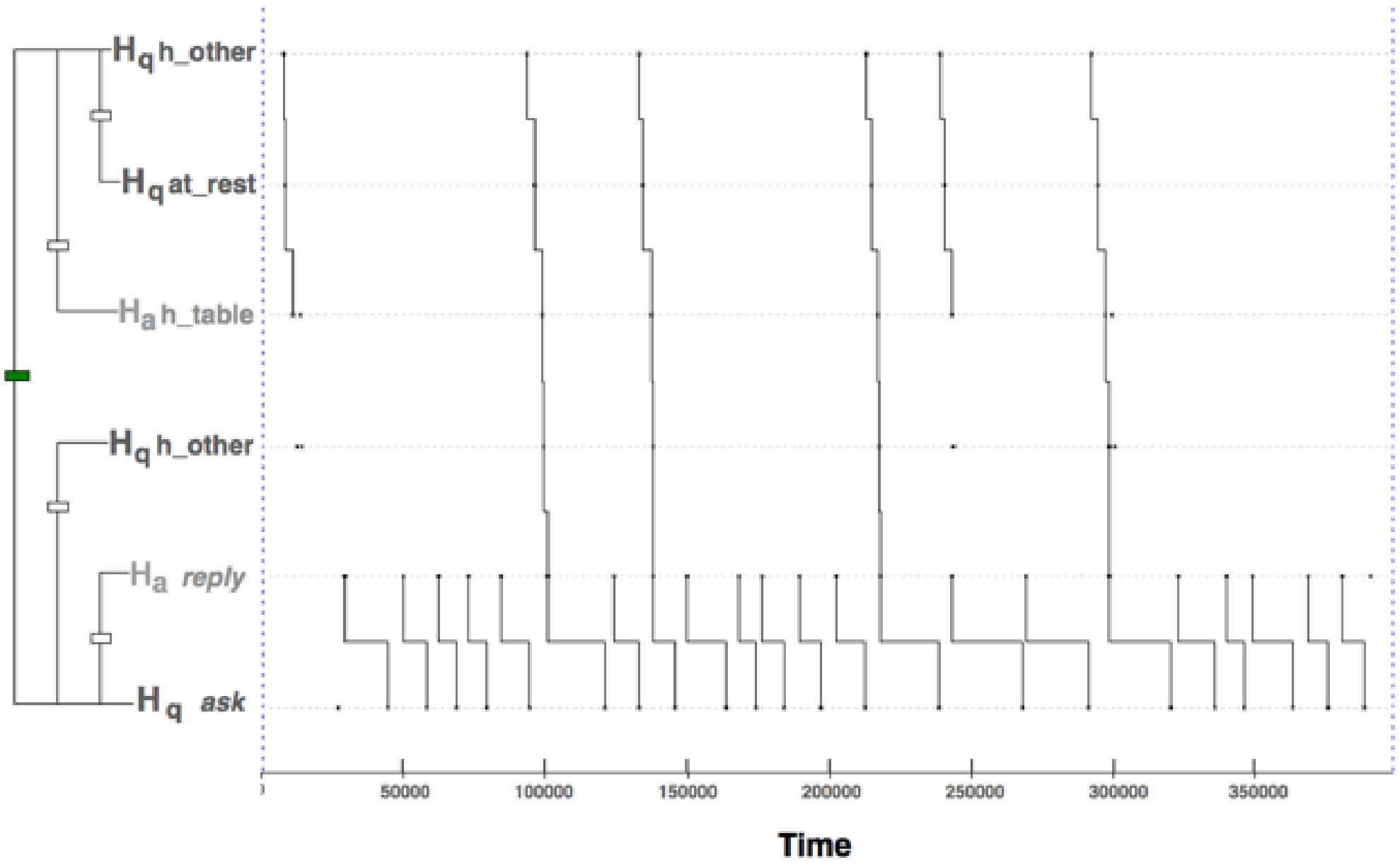

*Figure 16 - Example of common patterns seen in S1-as-interviewer condition in E2*

In text format, the tree is the following: ((( h2 h_partner, h2 at_rest ) h1 h_table )( h1,b,h_partner ( h1,b,reply,inf h2,b,ask,inf ))). Compare to human-human condition in Figure 15. (Timescale is in frames @30fps = 1 minute. For legend see Table 6.)

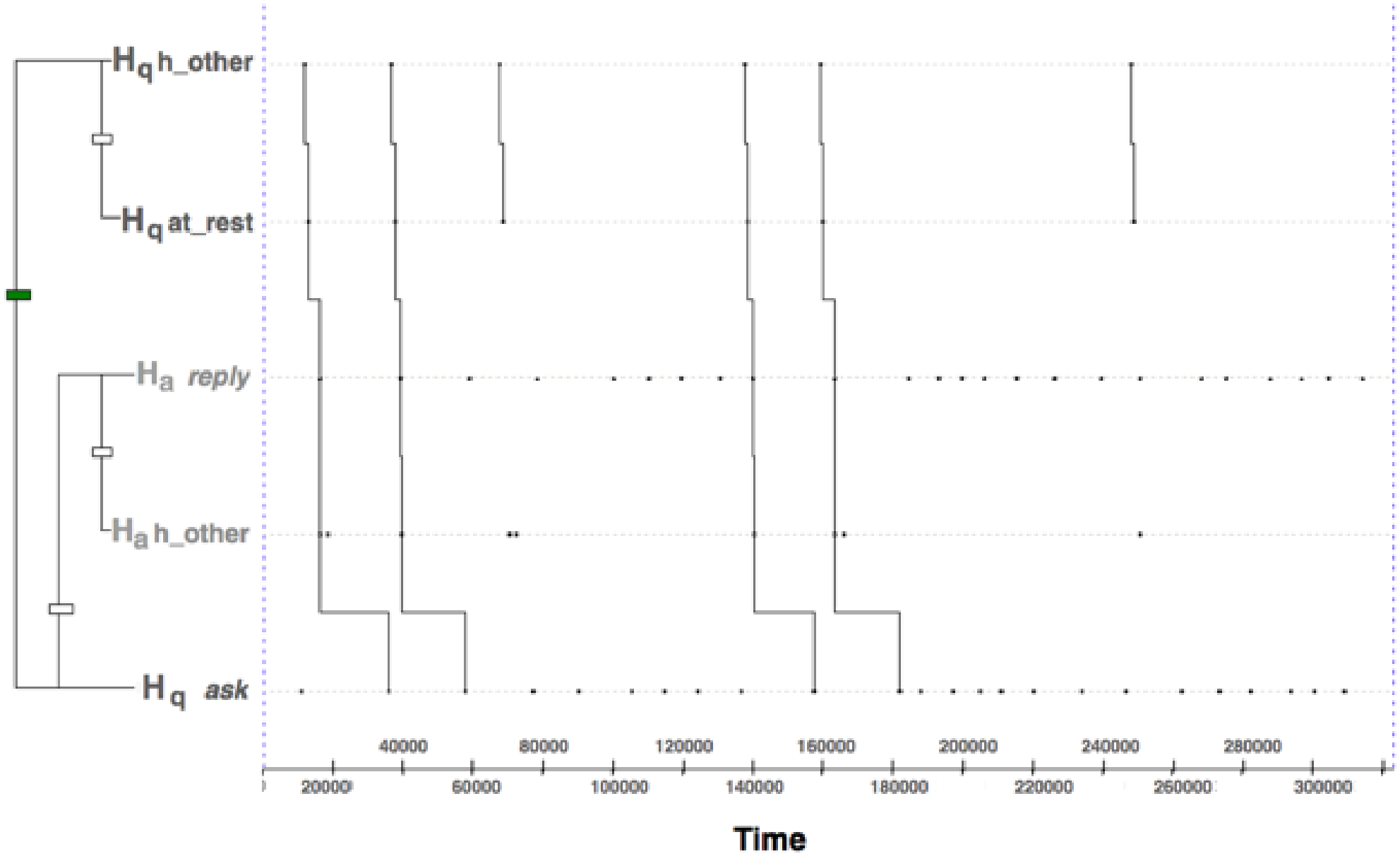

*Figure 17 - Example of common patterns seen in human-interviews-S1 condition in E2*

In text format, the tree is the following: (( h2 h_partner, h2 at_rest )(( h1 reply, h1 h_partner ) h2 ask )). *Compare to human-human condition in Figure 20.* (Timescale is in frames @30fps = 3.7 minutes. For legend see Table 6.)

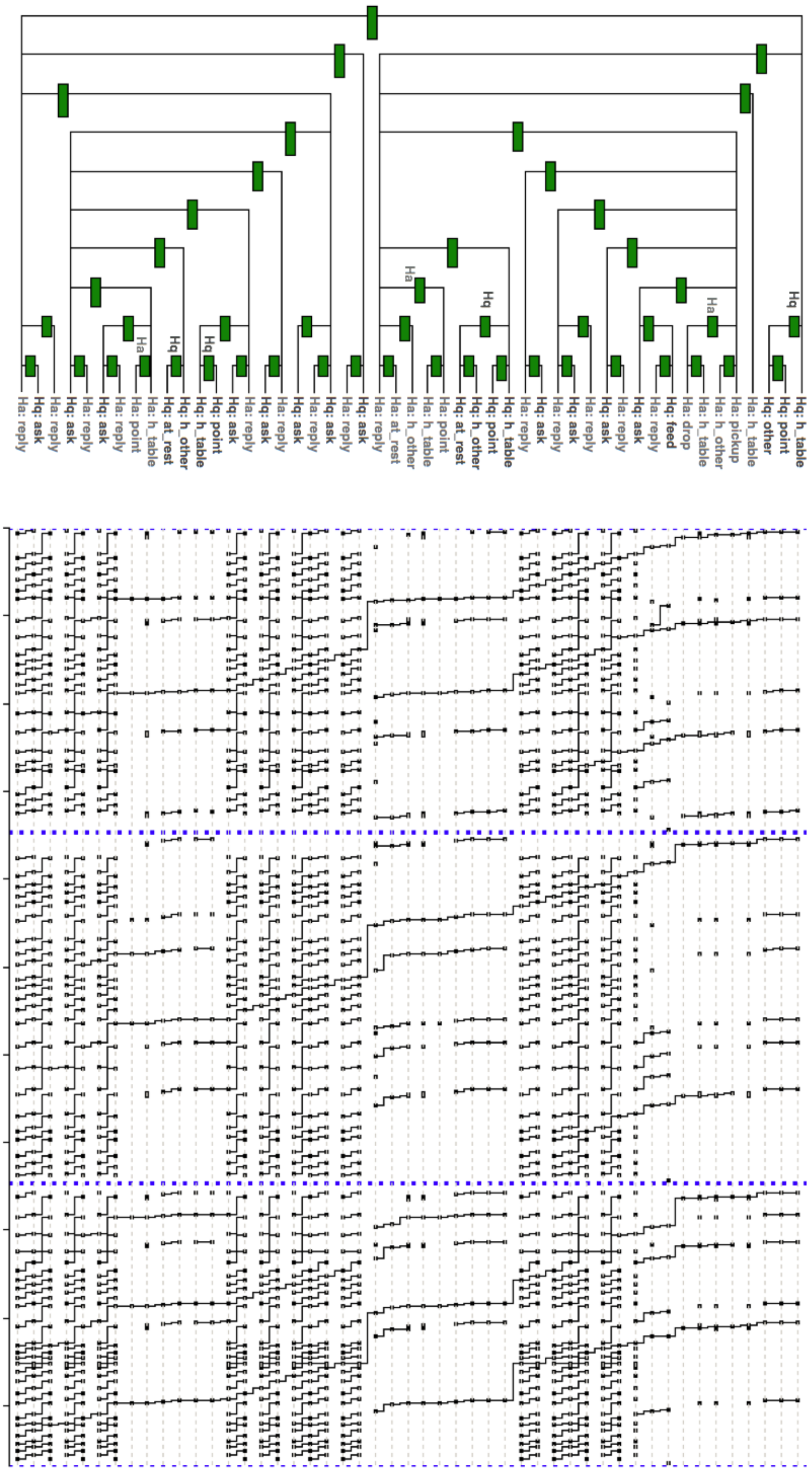

*Figure 18 - Non-overlapping patterns found in all conditions*

> This plot shows a non-overlapping hierarchical pattern found in all three conditions made up from 49 events (leaf nodes) occurring in the same order with significantly similar intervals between events. No two or more non-overlapping patterns were found to cover more of the total combined time, or 77% of the total combined time of all three conditions. The overall pattern and all its sub-patterns are significant at the $p < 0.005$ level. (Timescale is in frames; @30fps. For legend see Table 6.)

As the humans participating in these experiments are not trained actors, and did not follow a script, other than limiting their vocabulary and action repertoire as described above, there is good reason to believe that the skills thus learned by S1 generalize to a wider audience. However, the aim of this work was not specifically to build a control system that can interact with people but rather to create a system with a demonstrated ability to learn complicated tasks – human multimodal interaction of course being an excellent example of a domain with great complexity and variety in input, and one that is highly temporal, as in fact the vast majority of all real-world tasks are.

The variety in inputs demonstrated to handle goes a long way towards that goal, as clearly seen when we inserted a constraint on the duration of the interview. In the latter case, S1 observed the interviewer interrupting the interviewee towards the end of the interview, as the time limit was about to be reached. When S1 was put in the role of the interviewer, the human interviewee, on purpose, took more time than usual answering the first questions (a situation S1 had never seen before), and this triggered an interrupt from S1, as it predicted the overshooting of the interview's deadline and therefore the failure of its goals.

## 6.4 Summary of Results

The results from the two experiments E1 and E2 show without doubt that the AREA-based S1 system correctly acquired and mastered correct usage of all communication methods used by the human interviewer and interviewee in the human-human condition when conversing about the recycling of the various objects' materials. The complete absence of errors in S1's behaviors, after the observation periods in both conditions, demonstrate that very reliable models have been acquired, and that these form a hierarchy spanning at least two orders of magnitude in time. These models correctly represent the generalized relationships of a non-trivial number of entities, knowledge which had not been provided by the system beforehand by its designers and was acquired autonomously, given the bootstrap seed initially provided.

We set out to create a system that can learn, in a goal-directed way, how to achieve certain end-goals through generalized, autonomously acquired models that capture real cause-and-effect in the observed phenomena, providing a foundation for acquiring complex tasks through observation and reasoning. Our choice of a human interview to evaluate the system was not made on the grounds that we wanted to produce a highly human-like virtual interviewer, but was rather made purely on the grounds that it happens to meet a number of the key requirements that we have set ourselves for this work, and thus provided us a milestone of significant complexity that would test our work along several dimensions.

The tasks in E1 and E2 require S1 to learn and abstract temporal sequences of continuous events (utterances and multimodal behavior), as well as logical sequences and relationships (word sequences in sentences, meaning of words and gestures) between a number of observed data. These were acquired through a method of generalization using induction, abduction and deduction, allowing S1 to respond in real-time situations that differ from what it has seen before (the humans were not trained actors and did not repeat exactly any of their actions in any of the scenarios).

As all representations in AERA are domain-independent, these results are highly encouraging, implying that S1 and its successors can be applied to a vast number of tasks and environments, taking a notable step up the hierarchy illustrated in Figure 1.

# 7 Discussion & Related work

Most prior work on generality and life-long self-improvement differs considerably from ours in many respects, both theoretically and methodologically. Two main research paths in prior work can be discerned that can meaningfully be contrasted with ours. The first is an algorithmic approach to recursive self-improvement (cf. AIXI by Hutter 2005 and Gödel Machines by Schmidhuber 2006), based primarily on the tools and methodological stance of theoretical computer science; the second is work on cognitive architectures (cf. Laird 2012, Franklin 2013, Wang 2011, 2006), which shares our aim of designing holistic, complete systems, and is closer to implementation. Of these, Wang's stands out for taking a strong experience-grounded approach that we would classify as constructivist-centric, and his is the only work that obviously shares our goal of transversal life-long adaptation under an assumption of incomplete knowledge and limited resources (Wang 2006). The literature following these paths, especially that on cognitive architectures, counts a vast number of papers, ideas, proposals, and systems, but we can only fit a few select representatives. Here we will look at the work already cited in the order mentioned.

The Gödel Machine (GM) is a proposal to a general approach for imparting any system with recursive self-improvement and, as AIXI, remains to be implemented. Unlike AIXI, however, it *is* computable, and some research has suggested steps toward an actual implementation (Steunebrink & Schmidhuber 2012). However, as of yet, no such implementation does exist. A GM is composed of two main parts: an initial *solver* – a program that solves tasks in an environment – and a *searcher* – a program that searches for an improvement (a rewrite), proves its optimality, and subsequently applies it. The theory specifies the conditions under which such self-rewrites are allowed, but leaves vast problematic areas unattended. For example, designing an initial solver and its utility function that are general enough to handle non-trivial environments and tasks remains unaddressed. Moreover, designing the initial proof searcher presents an even bigger obstacle: generating suitable rewrite candidates requires either very powerful axioms (and thus tremendous foresight from the programmer) or a search that is probably so expensive as to be intractable. Indeed, the original algorithm (Schmidhuber 2006) proposes that the searcher enumerate theorems through Universal Search until it sees one which happens to contain a positive statement about self-modification. Assuming again that the search could be implemented in some practical way, we would then have to prove that a rewrite candidate will be both beneficial (with respect to the initial utility function) and better than continuing the search for other rewrites. Note that the proof will have to show that all cumulative future rewrites remain provably beneficial with respect to the initial utility function, which is extremely hard indeed. The difficulty of these fundamental remaining issues is compounded by the fact that the GM is too abstract to give – and therefore to exploit - any hints about the target of rewriting, i.e. the system itself, the very locus of construction, measurement and, eventually, improvement.

Our approach to recursive self-improvement is very different: We specify and constrain the system's architecture so as to *avoid* the aforementioned intractable issues. In our approach the "utility" of an improvement is not sanctioned by computationally expensive search and formal proofs but is sanctioned instead by observing an improvement's effect(s) on the performance of the system in its

environment – i.e. based on the system's experience over time. The spirit of this approach is more in line with other work of Schmidhuber et al. (1997), although AERA is designed to work in *real-world*, *real-time* scenarios. In AERA, finding improvements is this: When a prediction *fails*, or a goal *succeeds unexpectedly*, the system models the newly observed causal relationships and *tries them out*. Compared to the challenge of finding rewrites in the GM, this solution is trivial in terms of both computational cost and implementability. Unlike the GM, our system doesn't need shielding against destructive rewriting. In the AER approach, a system rewrites itself by continually adding and removing very small parts, not by modifying itself wholesale in successive, one-shot, global operations. Adding is always allowed because new code (models) is given an initial reliability that is very likely to be lower than that of established code and thus will not significantly change the operation of the system before enough positive evidences allow so; removing is also always allowed because (a) the necessity of doing so has been ascertained from having experienced some bad resulting behavior, and (b) it leads to an immediate replacement of the faulty part with several better candidates. The upshot is, we don't have to find rewrites: They are induced from observed facts, under the intrinsic necessity of improving the reliability of goals and predictions. Neither do we have to prove whether a rewrite is optimal or not: We know that the future will eliminate bad performers, so only the best available – not the best *possible* – will survive. Of course, such a pragmatic approach faces a potential pitfall: Shall the system have insufficient inputs (for example, in case the environment is too deceptive or adversarial), its rewrites would be useless. This is one reason why we have proposed a need for curiosity in our system, operationalized as the proactive improvement of a system's own knowledge (Steunebrink et al. 2013).

The GM framework arguably has a wider scope than our work is intended to have, and could possibly (in theory, at least) be applied to any system. However, it is not the generality of the *approach* that we are interested in, but rather the generality (and feasibility!) of the *system*. Because of the vast simplifications in algorithmic approaches, in that they for instance do not take into account system and resource constraints, it can ultimately be argued – without too much effort – that existing AGI work starting at the purely algorithmic end addresses a rather different problem than we do – so much so as to hardly be comparable.

The constraints of time and energy affect *all* intelligences, natural and artificial (cf. Thórisson 2013). Adaptation to constraints is key for survival and is one of the reasons why intelligent control has been deemed necessary by evolution and system engineers alike. Yet this fundamental property, adaptation, remains to be addressed in most of the work on cognitive architectures claiming biological inspiration or aiming to explain natural intelligence. Two frequently referenced architectures that aim for comprehensiveness over universality, SOAR (Laird 2012) and LIDA (Franklin et al. 2013), provide a case in point. In SOAR only one sub-state can be considered at a time, operators are not applied in parallel, and time-stamps on knowledge cannot be reasoned upon. Other critical design limitations are e.g. truth being axiomatic, and the system’s operation not being considered a target for reasoning, meaning that SOAR-based systems cannot support operational reflectivity. All these design choices strongly indicate that the self-management, and therefore the self-adaptation, of the cognitive load when facing limited available resources and knowledge was not on the research agenda. It is unclear how systems based on this architecture can improve over time in any goal-directed way – or why they even should. LIDA does not answer this question either, as details remain to be provided on its control strategy to explain how the system does actually allocate computational resources to the several types and instances of small code fragments, called codelets, that implement the LIDA cognitive cycle, knowing moreover that cognition in LIDA results from the parallel execution of several such cycles. Even where multi-threading is claimed to support the execution of the two cognitive architectures above, the available descriptions fall short of addressing how the allocation of threads and memory impacts the operation of the system – that is, how it affects cognition – and conversely, how the system learns to adapt its load to the resources at its disposal – i.e. how cognition impacts the allocation of resources and improves it. In contrast, our approach places the inter-dependency of cognition and adaptation at the very heart of the design of control systems.

Of all prior work, Wang's non-axiomatic reasoning approach (Wang 2006) shares with us by far the largest set of goals and assumptions. Wang has been virtually alone in championing the assumption of incomplete knowledge and limited resources (AIKR, Wang 2011), which we have adopted. His NARS (non-axiomatic reasoning system) is based on NAL (non-axiomatic logic), a novel term logic designed to handle uncertainty and growing experience. Its nine levels of expressivity, some of which share fundamental traits with our own logic (for example, equivalents of our likelihood, experience, reliability and success rate), are described in detail in Wang (2006).

NARS is a work in progress – an implementation already exists including NAL levels up to six. Even though NAL is quite developed, defining a richer set of inferences than we have implemented so far in AERA’s S1 (e.g. analogy making, detection of equivalences, similarity), it does not include representations of time at its very core – time is only handled at the periphery (NAL level 7), which makes it difficult to control real-time systems. Equations in NARS are treated as first-class citizens, encoded as networked independent terms. While its design is in some ways quite elegant it may be overly complicated for both learning and exploitation. However, compared to NARS AERA is limited in other ways – for example, its abstraction capabilities are less developed than NARS’s.

These differences aside, and from a broader perspective, NARS’s take on cognition is very similar to ours. As in AERA, cognition in NARS can be considered a by-product of allocating resources to cognitive elementary tasks, coupling cognition and resource allocation in a mutual and continual feedback loop. Due to NARS's potential to be (fully) implemented in the coming months or years, the differences and related underlying principles of our respective approaches can potentially be compared experimentally, a much faster path than relying solely on theoretical debate. Such comparison would likely results in improvements to both approaches.

Even if our respective agendas differ on their over-arching goals - NARS is intended to capture features of human intelligence, whereas we want to build intelligent controllers for machinery – our principles are compatible and our systems, to some extent, exhibit common properties. This indicates that perhaps an adherence to AIKR is at least as important as other considerations and methodological issues when establishing the requirements for the design of cognitive architectures.

# 8 Conclusion & Future Work

Generality and recursive self-improvement are key capabilities for systems intended to continually maintain their effective delivering of value to their owners, despite unforeseen and possibly adversarial situations. These are fundamental challenges for artificial general intelligence to which we, going beyond prior attempts in this direction, present experimental results and demonstrable solutions.

We have demonstrated an implemented architecture that can learn autonomously many things in parallel, at multiple time scales. The results show that AERA system S1 can learn complex multi-dimensional tasks from observation, while provided only with a small ontology, a few drives (high-level goals), and a few initial models, from which it can autonomously bootstrap its own development. This is initial evidence that our CAIM methodology, based on the principles of AER, is a way for escaping the constraints of current computer science and engineering methodologies. Human dia-logue is an excellent example of the kinds of complex tasks current systems are incapable of handling autonomously. The fact that no difference of any importance can be seen in the performance between S1 and the humans in simulated face-to-face interview is an indication that the resulting architecture holds significant potential for further advances.

That being said, in its current incarnation AERA is entirely dependent on observation, as learning is exclusively triggered by unexpected goal achievement, or a prediction that turns out to be wrong – i.e. by surprise. This limits the acquisition of knowledge to phenomena that are directly observable – hidden causation is difficult for the current system to figure out, as are other kinds of inexplicit relations (similarity, equivalence, etc.). Another functionality the system lacks is some form of curiosity, as it is not able to produce hypotheses and conduct related experiments to eventually learn proactively, for the system to be intrinsically motivated to fill the gaps in its (inevitably limited) knowledge. One form of this is “motor babbling”, which we have experimented with in the prototype to bootstrap learning. So far this has relied on ad-hoc hand-crafted goals injected into the system at strategic times by the designers, and therefore this technique was not allowed to be used in the S1 agent presented here. The question of how to generate such goals in a principled way automatically is currently being investigated. In an extension of the AIKR assumption, we have argued that curiosity results from the need to overcome the limitations imposed by the scarcity of inputs (Steunebrink et al. 2013). We plan to expand the types of programs to implement a richer set of inferences from which curious behaviors can be devised and planned, whenever the system has resources to spare. Interestingly, when improving the capabilities of our architecture thus far we have strictly proceeded by addition rather than by modification (“hacking”): We conjecture that our methods are amenable to significant and tractable improvement, but far from claiming its theoretical universality we bet instead on its practical scalability, which leads us to our last and final remark.

To support the claim of achieving bounded self-improvement *in general*, one has to demonstrate a fundamental property namely, scalability. This work is still too recent for making an assessment to support such a claim. One of the main directions of our planned near-future work is thus set toward building more prototypes to provide more evidence for the generality of our system and, for its scalability. This endeavor, no doubt, will make use of the advanced self-modeling and self-control facilities that have been already implemented. In particular, we will leverage the self-compilation capacity mentioned in section 5.3, as one of the main thrusts toward scaling up.

### Acknowledgments

This work was supported by the European Project HUMANOBS – Humanoids that Learn Socio- Communicative Skills By Observation (FP7 STREP – Cognitive Robotics - Grant number 231453), and by research grants from RANNIS, Iceland. We are grateful to Th. Bryndis Thorisdottir for the design of the interview, Gunnar S. Valgardsson and Hrafn Th. Thorisson for the design of the avatars. Last but not least, we thank Pei Wang for valuable and constructive discussions.